\pdfoutput=1
\documentclass[conference]{IEEEtran}

\usepackage{times}

\usepackage[numbers]{natbib}
\usepackage{multicol}
\usepackage[bookmarks=true]{hyperref}
\usepackage{graphicx}
\usepackage{amsmath}
\usepackage{amssymb}
\usepackage{xcolor}
\usepackage{subfig}
\usepackage{algpseudocode,algorithm,algorithmicx}
\usepackage{amsthm}
\usepackage{tabularx}
\usepackage{multirow}
\usepackage{wrapfig}

\usepackage[font=scriptsize,labelfont=bf]{caption}

\newtheorem{theorem}{Theorem}
\newtheorem*{remark}{Remark}
\newtheorem{assumption}{Assumption}
\graphicspath{{/figurs/}}  

\DeclareMathOperator*{\argmax}{arg\,max}
\newcommand{\prob}[1]{\ensuremath{\mathbb{P}({#1})}}
\newcommand{\expt}[2]{\ensuremath{\underset{{#1}}{\mathbb E}{\{{#2}\}}}}
\newcommand{\maxim}[2]{\ensuremath{\underset{#2}{#1}}}
\newcommand{\ubrace}[2]{\underbrace{#1}_\text{#2}}

\begin{document}
\title{Online POMDP Planning via Simplification}



\author{Ori Sztyglic$^1$ and Vadim Indelman$^2$ \\
	$^1$Department of Computer Science \quad $^2$Department of Aerospace Engineering\\
	Technion - Israel Institute of Technology, Haifa 32000, Israel\\
	\small{\texttt{ori.sztyglic@gmail.com, vadim.indelman@technion.ac.il}}
}


%

\maketitle

\begin{abstract}
In this paper, we consider online planning in partially observable domains. Solving the corresponding POMDP problem is a very challenging task, particularly in an online setting. 
Our key contribution is a novel algorithmic approach, \textit{Simplified Information Theoretic Belief Space Planning} (SITH-BSP), which aims to speed-up POMDP planning considering belief-dependent rewards, without compromising on the solution's accuracy. We do so by mathematically relating the simplified elements of the problem to the corresponding counterparts of the original problem. Specifically, we focus on belief simplification and use it to formulate bounds on the corresponding original belief-dependent rewards.
These bounds in turn are used to perform branch pruning over the belief tree, in the process of calculating the optimal policy.  We further introduce the notion of adaptive simplification, while re-using calculations between different simplification levels, and exploit  it to prune, at each level in the belief tree, all branches but one. 
Therefore, our approach is guaranteed to find the optimal solution of the original problem but with substantial speedup. As a second key contribution, we derive novel analytical bounds for differential entropy, considering a sampling-based belief representation, which we believe are of interest on their own. We validate our approach in simulation using these bounds and where simplification corresponds to reducing the number of samples, exhibiting a significant computational speedup while yielding the optimal solution.

\end{abstract}

\IEEEpeerreviewmaketitle

\section{Introduction}

In the world of autonomous agents operating in an uncertain environment, a Partially Observable Markov Decision Process (POMDP) provides a principled mathematical framework for planning under uncertainty. Solving a POMDP is proven to be PSPACE-Complete \cite{Papadimitriou87math} giving rise to many algorithms trying to approach the optimal solution without having to solve the entire problem. This difficulty is more keenly felt when considering an online-setting of such autonomous tasks, i.e.~when an  agent has few seconds at most in every time step till it  needs to execute the action it deems to be 'optimal'.

\begin{figure}[t]
	\centering
	\includegraphics[width=\columnwidth]{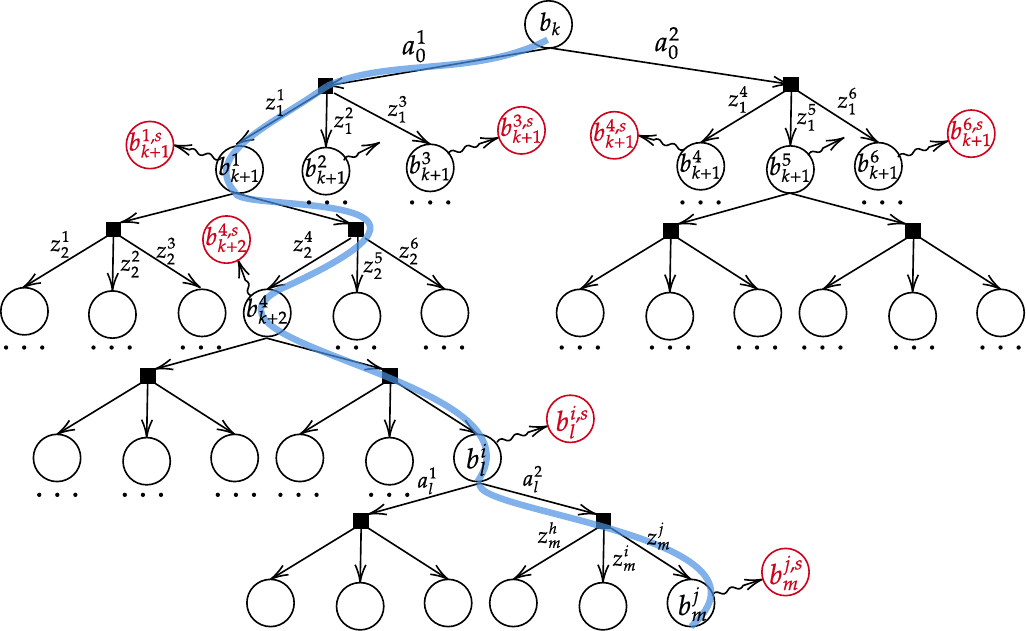}
	\caption{In-place belief tree simplification:  For each belief node $b$ in a given belief tree a simplified belief $b^s$ is calculated and used to formulate bounds over the corresponding belief-dependent reward. These bounds are then used to prune branches while calculating the optimal policy.
	}
	\label{fig:inplace_simp}
\end{figure}

In a partially observable online setting, the agent maintains a \emph{belief}, a posterior distribution over the state of interest,  given  actions and observations history the agent has executed and gathered so far.
At each planning session, given this belief the agent determines the optimal policy (or action sequence)  by constructing and traversing a belief-tree, as illustrated in Fig.~\ref{fig:inplace_simp}, which  models how the POMDP can evolve into the future considering some finite horizon of time steps. When constructing the tree at planning time, the tree branches when taking an action and again when acquiring an observation. Since the agent has some purpose or goal, the optimal path down the tree is determined by some reward or cost each belief node in the tree induces.

This setting presents two main difficulties. The first is the \textit{curse of dimensionality}. The size of the state space grows exponentially with the number of state variables and correspondingly so does the belief. The second is the \textit{curse of history}. Planning into the future requires to build the belief tree,  which grows exponentially with the action and observation spaces. These two problems gave rise to many works  trying to reduce computation time when solving the POMDP such as \cite{Silver10nips} and \cite{Smith04uai}. In recent years many of these works began to address the POMDP setting considering more 'real-world' and complex settings. One example is considering continuous or huge observation and action spaces e.g.~\cite{Sunberg18icaps, Garg19rss, Lim20ijcai}. Another example is papers considering information-theoretic rewards such as \cite{Fischer20icml}. The latter requires to maintain a posterior distribution over the state (the belief). On the up side, these information gathering rewards can be extremely beneficial, however they can also be very costly (computation speaking) and accounting for each node in the belief tree the problem quickly becomes even more intractable.
	
In this work, motivated by real world problems, we suggest a notion called \textit{Simplification}. The idea is when given some 'hard-to-solve' problem, bypass it by simplifying elements of the original problem and mathematically relate them to the corresponding counterparts of the original problem. This will ultimately lead to the optimal solution. Though the notion of simplification is not new in the field of POMDP and belief space planning, to the best of our knowledge this is the first time simplification is used as a way to initialize and update bounds over the tree branches. 

Our suggested approach is general: given some belief tree constructed for the current planning step we consider simplification of one or more of the POMDP elements such as the reward, observation and motion models. We consider rewards over the belief instead of the states (like many POMDP solvers do). This setting is more general and can easily be reduced to the well known problem of planning over state rewards. Our approach admits information theoretic rewards, which are particularly important in robotics etc. and Lipschitz Continuous rewards for control oriented behavior. Moreover, we consider sampling-based methods for belief representation.

As our second key contribution, we derive novel analytical bounds for the differential entropy reward function, which is not Lipschitz Continuous. These fast calculating bounds rely on the simplified version of the problem corresponding to taking a subset of samples representing the belief, and we show they converge to the costly differential entropy reward function of the original problem. 
Importantly, these bounds can be made sufficiently tight to allow elimination of branches in the belief tree.

Finally we present an adaptive way to apply our simplification approach, avoiding the need to predetermine the simplification level, and letting the algorithm to identify an appropriate simplification level by itself. Crucially, we show this can be done while re-using calculations between different simplification levels.

To summarize, our main contributions are as follows: (a) For the setting of online POMDP planning we introduce simplification as a paradigm to gain speedup using branch and bound over the belief tree. (b) We derive novel analytical bounds on approximation of the differential entropy which depend on the chosen simplification. (c) Assuming LC reward functions, we derive novel bounds on the objective function that depend on the chosen simplification. (d) We show the mentioned bounds converge to the original rewards and thus the objective as well. (e) We develop an adaptive simplification scheme, where different simplification levels  can be applied to the problem in a dynamic online manner. Further we show it can be done while re-using calculations when moving from one simplification level to another, saving computation time.

\section{Related Work}
Although POMDPs are a general method for planning and decision making under uncertainty, they are computationally difficult to solve. State of the art POMDP solvers are roughly divided into to two categories, offline and online solvers. In this work we focus on the online setting. 

POMCP by \citet{Silver10nips} uses Monte-Carlo Tree Search (MCTS) for POMDP solving in an online setting. They use particle filtering and the UCB1 algorithm to guide their action selection. Unlike us they do not make use of branch-and-bound methods to prune the tree and their rewards are a function of the state instead of the belief.
 	
\citet{Sunberg18icaps} extend POMCP to POMCPOW and PFT-DPW, aiming to handle large observation and action spaces (continuous or discrete). Furthermore, PFT-DPW practices rewards over the belief. However they do not consider simplified elements of the original problem like we do and they make no use of bounds for pruning tree branches.
 		
\citet{Lim20ijcai} provide mathematical analysis and guarantees for the convergences of such MTCS algorithms. They consider two different settings of weighted and unweighted particles. For the continuous observation setting they use the weighted approach and the number of observations made at each tree node is the same as the number of particles used to represent the belief. Combined with the fact they carry out breadth first search for the tree expansion, their approach is very time consuming. However their analysis relies on this formulation. We too consider weighted particles, and since their tree is very different than other approaches we show our approach can be also applied to their constructed tree.
 		
\citet{Fischer20icml} suggest IPFT, an extension of PFT-DPW to information theoretic rewards. They focus on the differential entropy and show how they may calculate it efficiently using kernel density estimation and the particles representing the belief. Our paper differs from theirs in the way we address the entropy calculation and the novel, analytical bounds we propose on the differential entropy approximation.

Heuristic Search Value Iteration (HSVI) by \citet{Smith04uai, Smith05uai} is an early seminal work that elevated bounds over the belief tree to their stature in the context of POMDP planning. Their bounds guide their heuristics and by doing so, prune sub-optimal branches. The bounds are an upper and lower approximations to the true value function. However the lower bound is costly to maintain and initialization of the bounds places them far away from one another. The bounds serve as a stopping criteria, meaning when they are close enough they stop and return the current policy. This means that in an online setting after time out the bounds may not be close enough and the output may be sub-optimal. Our interpretation and usage of the bounds is different, as discussed in the sequel.
 	
 	
 	
 	
 	
One of the state-of-the-art POMDP online solvers is DESPOT by \citet{Ye17jair} and following work DESPOT-$\alpha$ by \citet{Garg19rss}. DESPOT follow the heuristics introduced by Smith et al. (HSVI), i.e. they use Bellman updates on their bounds in order for them to guide the actions selection. As previous approaches, rewards are over the states, not the belief.
 	
\citet{Fehr18nips} use Lipschitz continuous (LC) properties of the reward model. They too follow the implementation of HSVI. They show the objective function and the bounds are LC and the Bellman operator preserves these LC properties of the bounds.
We too consider LC properties, however our approach does not assume LC rewards in general and can be applied to rewards which are not LC.  Moreover, we propose to mathematically relate the simplified elements of the POMDP to the corresponding counterparts of the original problem,  
 and show this is possible if we can relate the two problems via appropriate bounds. One possible formulation of these bounds is using LC properties; however, we show how for the non-LC case we can derive novel analytical bounds depending on the simplified elements of the problem.

 	
\citet{Hoerger19isrr} are more closely related in concept to our work. They plan over the POMDP in an online setting and preform sampling from  increasing approximation levels of the POMDP's motion and observation models. In their setting the levels are correlated, i.e. they enforce same action sequence for all the approximation levels. They use UCB1 to guide their action selection (as done in POMCP \cite{Silver10nips}) and do not make use of branch-and-bound methods. Finally, they consider rewards over state and not the belief. In contrast, our approach utilizes simplified elements of the problem to bound the original problem, while considering belief-dependent rewards. Furthermore, we use an online adaptive paradigm for the level of approximation (simplification).
 	
 	
\citet{Elimelech18ijrr_submitted} formulate the notion of simplification in a general manner but their identified and implemented simplification is restricted to the Gaussian case and maximum likelihood assumption. In contrast, our approach is formulated for general belief distributions; furthermore,  we use (like most mentioned approaches) particle filtering for planning and decision making, 
in an online setting.

\section{Background}

\subsection{POMDP}
We model the Partially Observable Markov Decision Process (POMDP) for the finite horizon case, as a 7-tuple: $M = (\mathcal{X},\mathcal{A},T,r,\mathcal{Z},\mathcal{O}, b_0)$, 
where $\mathcal{X}, \mathcal{A}$ and $\mathcal{Z}$ are the state, action and observation spaces, respectively.  $T$ is the probabilistic transition model and expresses the probability to transit from state $x \in \mathcal{X}$ to state $x'\in \mathcal{X}$ by executing action $a\in \mathcal{A}$, i.e. $T(x,a,x') \triangleq \prob{x'\mid x,a}$. $\mathcal{O}$ is the observation model expressing the measurement likelihood for some given state, i.e.~$\mathcal{O}(x,z) \triangleq \prob{z\mid x}$, with $z \in \mathcal{Z}$. 
%
%
$b_0$ is the initial belief we have on the state at planning time.
The \textit{belief} is  a posterior distribution over the state given all actions and measurements so far. 
It can be updated recursively via Bayes rule as $b[x']=\eta \int \prob{z' \mid x'}\prob{x'\mid x,a} b[x]dx$, where $\eta$ is a normalization constant. 

Unlike many prominent works, in this paper we consider a belief-dependent reward function $r(b,a)$. Such a formulation is more general than typical state-dependent reward functions; in particular it allows one to use information theoretic costs such as (differential) entropy, information gain and mutual information, thereby reasoning about future posterior uncertainty within the decision making process.
%

We denote the posterior belief at planning time  $k$ as $b[x_k] \triangleq \prob{x_k \mid a_{0:k-1}, z_{1:k}}$.  Further, we denote  by $\pi_{k+j}$ a policy for time step $k+j$, i.e.~$\pi_{k+j}(b[x_{k+j}])$ determines the action $a_{k+j}$. Let  $\pi_{k+} \triangleq \pi_{k:k+L-1}$ represent a  sequence of policies for the entire planning horizon of $L$ steps that starts at time instant $k$.  To shorten notations, we shall also use in the sequel $\pi_{(k+j)+} \triangleq \pi_{k+j:k+L-1}$, as well as $b_{k+j} \triangleq b[x_{k+j}]$. We denote \emph{history} at time $k$ to be all actions taken and captured observations up to this point, $H_k \triangleq \{a_{0:k-1}, z_{1:k}\}$.



%
%
%
%
%
%
When solving a POMDP,  one is trying to find the optimal policy that maximizes the expected long term cumulative rewards, also known as the objective (value) function,
\begin{equation}\label{eq:objective}
	J(b_k,\pi_{k+})= 
	\expt{z_{k+1:k+L}}{\sum_{i=k}^{k+L-1}r(b_{i},\pi_i(b_i))+r(b_{k+L})},
\end{equation}
where 
$r(b_{k+L})$ is the terminal reward which is therefore not a function of action. 
%
We may also consider a more general reward structure of $r(b_i, b_{i-1}, a_i)$, which is required, for example, to support information-theoretic reward functions such as information gain, and a specific sampling based approximation of differential entropy \cite{Boers10fusion} that we shall use in Section \ref{bounding_entropy_section}. As earlier, action $a_i$ is determined by $\pi_i(b_i)$.

The optimal policy $\pi^{\star}_{k+} \triangleq \pi_{k:k+L-1}^{\star}$ and the corresponding objective function are given by 
\begin{equation}\label{eq:argmax_objective}
\pi_{k+}^{\star} = \argmax_{\pi_{k+}}{J(b_k,\pi_{k+})}, \ \ 
J^{\star}(b_k) = \max_{\pi_{k+}}{J(b_k,\pi_{k+})}.
\end{equation}
Further, the objective function \eqref{eq:objective} can be written recursively, 
\begin{equation}\label{eq:value_fun}
J(b_k,\pi_{k+}) \! = \!
r(b_k,a_k) +\!\! \expt{z_{k+1}}{\!J(b_{k+1},\pi_{(k+1)+})},
\end{equation}
which, in turn, leads to the Bellman optimality equation,
\begin{equation}\label{eq:bellman}
	J(b_k, \pi^{\star}_{k+}) =
	\max_{\pi_{k}} \{r(b_k,a_k)+\!\! \expt{z_{k+1}}{J(b_{k+1},\pi^{\star}_{(k+1)+})}\}.
\end{equation}


\subsection{Planning Using Reward Bounds}\label{branch-and-bound}
\begin{figure}[!ht]
	\centering
	\subfloat[ \label{fig:action_graph}]{\includegraphics[width=0.33\columnwidth]{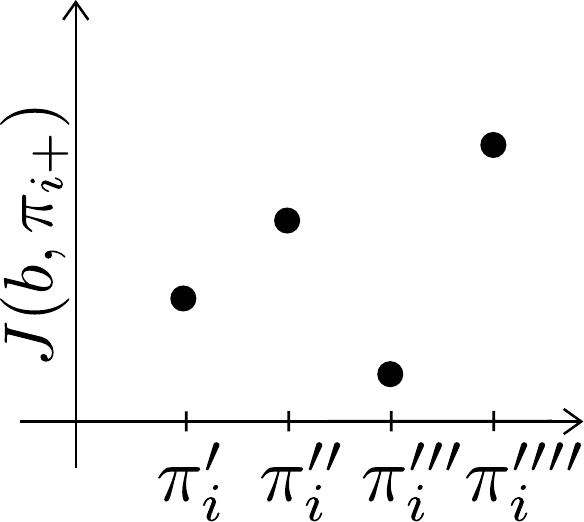}}
	\subfloat[ \label{fig:action_graph_s0}]{\includegraphics[width=0.33\columnwidth]{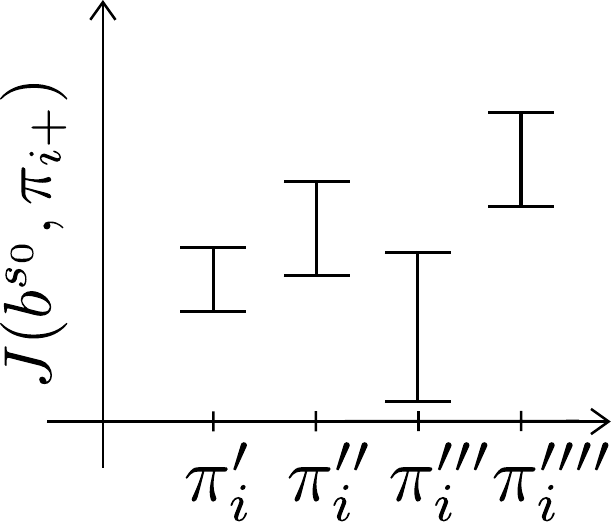}}
	\subfloat[ \label{fig:action_graph_s1}]{\includegraphics[width=0.33\columnwidth]{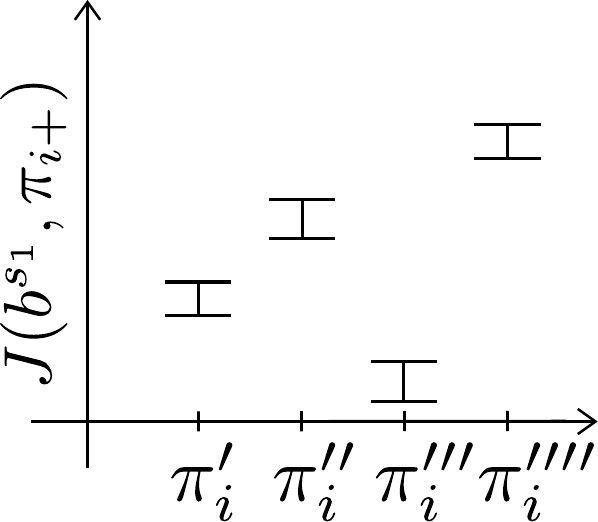}}
	\caption{Action elimination using bounds. \textbf{(a)} Objective for some belief $b$ and candidate policies; \textbf{(b)} Objective bounds given belief is simplified to level $s_0$; \textbf{(c)} Objective bounds given belief is simplified to level $s_1$.}
\end{figure}

In this paper we make use of pruning in order to speed up planning. Essentially, in the context of trees, pruning is done by eliminating some of the sibling branches (ideally all but one) using upper and lower bounds corresponding to each branch. To better understand, assume we are trying to prune one of two sibling subtrees $m',m''$ with corresponding bounds  $\{(\mathcal{LB}^{m'}, \mathcal{UB}^{m'}), (\mathcal{LB}^{m''}, \mathcal{UB}^{m''})\}$. We can prune $m'$ if the bounds hold $\mathcal{LB}^{m''} > \mathcal{UB}^{m'}$. E.g., in Fig.~\ref{fig:action_graph_s1} the lower bound of $\pi''''$ is higher than all other actions upper bounds.
For a more detailed explanation see the Appendix~\ref{appen:branch_and_bound}.

Although the use of bounds is extensive(\cite{Smith05uai, Fehr18nips, Silver10nips}), our use and interpretation of them is very different. E.g.~they do not serve as stopping criteria nor use to guide some heuristic.

\section{Our Approach}\label{approach}
\subsection{Simplification}\label{sec:simplification}
When speaking about simplification we refer to any sort of relaxation in the problem we currently try to solve. Recall the POMDP is a tuple $M$, composed out of elements. Each of these elements can be simplified in order to get an easier problem to solve. We denote $M^s$ as the POMDP tuple corresponding to the simplified problem.
How can another easier problem help solving a more difficult problem?
The simplified problem $M^s$ has its own solution $\pi^{s \star}$ since the simplification can induce a different belief tree.

While the simplified POMDP problem can be computationally easier to solve, there is no guarantee $\pi^{s \star}=\pi^{\star}$.  
However, calculations involved in solving the simplified problem $M^s$ can be mathematically related to the original problem $M$ and provide guarantees over its solution. Thus, we can avoid from calculating the full 'expensive' solution of $M$.




In this paper we consider a specific instantiation of this general simplification framework;  
namely, we suggest to simplify the reward model $r$. This results in a simplified POMDP $M^s=(\mathcal{X},\mathcal{A},T,r^s,\mathcal{Z},\mathcal{O}, b_0)$, which corresponds to the simplified objective function: 
\begin{equation}\label{eq:simplified_objective}
\begin{split}
&J^s(b_k,\pi_{k+})=
\!\!\!\! \expt{z_{k+1:k+L}}{\!\!\sum_{i=k}^{k+L-1}r^s(b_{i},\pi_i(b_i))+r^s(b_{k+L})},
\end{split}
\end{equation}
instead of \eqref{eq:objective}. Note, in this setting, simplification does not impact the distribution over which the expectation is taken. Thus, a given belief tree of the original problem $M$, corresponds also to the simplified problem $M^s$.  
We denote this setting as \textit{in-place simplification} (IPS) and we note here that this setting assumes the belief tree was built in some manner and it is given.

\begin{remark}
	In some approaches the reward model is involved in the construction of the belief tree. Accounting the tree construction is beyond the scope of this paper and we leave it to future research.
\end{remark}
Simplification of the reward model $r(b,a)$ can be realized in different ways. Generally speaking, one could consider simplifying the function $r(.,.)$ or/and feeding to the original reward model a simplified version of the belief. In this paper we focus on the latter, although the formulation presented in this section is general and thus applicable also to the former case. Specifically, we denote the simplified reward model as 
\begin{equation}\label{eq:reward_simplifcation}
r^s(b,a) \triangleq r(f(b), a) =  r(b^s, a),
\end{equation}
where  $f(\cdot)$ is a belief simplification operator, i.e.~$b^s=f(b)$. We discuss a specific such operator in Section \ref{simplification}.

Further, we consider the original reward model can be bounded using the simplified problem. Generally, these bounds can be of the form
\begin{equation}\label{eq:bound_reward_general}
	\mathbf{lb}(b^s,b,a) \leq r(b,a) \leq \mathbf{ub}(b^s,b, a), 
\end{equation}
where $\mathbf{lb}$ and $\mathbf{ub}$ are the corresponding lower and upper bounds, respectively. A key requirement is reduced computational complexity of these bounds compared to the complexity of the original reward. 


We note that, in particular, these bounds could directly involve the simplified reward model, as in
%
%
\begin{equation}\label{eq:bound_reward}
	\mathbf{lb}(r^s(b,a),  b^s, b, a) \leq r(b,a) \leq \mathbf{ub}(r^s(b,a), b^s, b, a), 
\end{equation}
as $r^s(b,a)$ is a function of $b^s$ and $a$ according to \eqref{eq:reward_simplifcation}.

Furthermore, our formulation can be extended straightforwardly  to support also information-theoretic rewards of the form $r(b_{i-1},b_{i})$, which involve two (consecutive)  beliefs $b_{i-1}$ and $b_i$, such as information gain. In such a case, the corresponding bounds would be
\small
\begin{equation}\label{eq:bounds_extended}
	\mathbf{lb}(b_{i-1}^s, b_i^s, b_{i-1}, b_{i}) \leq	 r(b_{i-1}, b_{i}) \leq  \mathbf{ub}(b_{i-1}^s, b_i^s, b_{i-1}, b_{i}). 
\end{equation}
\normalsize
In this section we formulate our approach considering the general form of the bounds  \eqref{eq:bound_reward_general}.  In Sections  \ref{LC_bounds_section} and 
\ref{bounding_entropy_section} we derive bounds of the form \eqref{eq:bound_reward} and \eqref{eq:bounds_extended}, respectively, considering common reward models.

%
%

%


What use do these reward bounds have? Consider a given belief tree of the original problem $M$. Instead of calculating the expensive reward $r(b,a)$ for each belief node $b$ in this belief tree, we first calculate the corresponding simplified belief $b^s$, as illustrated in Fig.~\ref{fig:inplace_simp}, and then formulate the bounds $\mathbf{lb}$ and $\mathbf{ub}$ from \eqref{eq:bound_reward_general}. 



Moreover, we can now traverse the belief tree from the leafs upwards and calculate recursively bounds on the objective (value) function at each node $b_{i}$ via Bellman equation \eqref{eq:value_fun} as described below for $i\in[k,k+L-1]$.
\begin{equation}\label{eq:recursive_simp_bounds}
	\begin{split}
		&\mathcal{UB}(b_i, \pi_{i+}) \!=\! \mathbf{ub}(b^s_i, b_i, a)) \!+ \!\!\! \expt{z_{i+1}}{\mathcal{UB}(b_{i+1},\pi_{(i+1)+})}\\
		&\mathcal{LB}(b_{i}, \pi_{i+}) \!=\! \mathbf{lb}(b^s_i, b_i, a) \!+ \!\!\! \expt{z_{i+1}}{\mathcal{LB}(b_{i+1}, \pi_{(i+1)+})},
	\end{split}
\end{equation}
with $\pi_{i+} = \{ \pi_i, \pi_{(i+1)+}\}$ and $a=\pi_i(b_i)\in \mathcal{A}$, and where the expectation is taken with respect to $\prob{\cdot \mid H_{i},a}$, and the bounds are initialized at the terminal rewards ($L$th time step in the planning horizon) as $\mathcal{LB}(b_{k+L})=\mathbf{lb}(r^s(b_{k+L}))$ and $\mathcal{UB}(b_{k+L})=\mathbf{ub}(r^s(b_{k+L}))$. This recursive procedure is common and practiced in many works (e.g. \cite{Ye17jair}), yet a key difference is that our bounds \eqref{eq:bound_reward} are obtained by relating the simplified POMDP elements to the original problem.


Eq.~\eqref{eq:recursive_simp_bounds} is a recursive update considering \emph{some} trajectory down the belief tree determined by some policy $\pi_{i+}$. In contrast, we now consider upper and lower bounds for the  optimal policy $\pi^{\star}_{i+}$. We denote these bounds as
\begin{equation}\label{eq:BoundsOpt}
	\mathcal{UB}^{\star}(b_i) \triangleq \mathcal{UB}(b_i, \pi^{\star}_{i+}), \quad \mathcal{LB}^{\star}(b_i) \triangleq \mathcal{LB}(b_i, \pi^{\star}_{i+}).
\end{equation} 
Updating the bounds \eqref{eq:BoundsOpt} is done recursively in two steps. First by considering the expansion of the already-calculated bounds $\mathcal{UB}^{\star}(b_{i+1})$ and $\mathcal{LB}^{\star}(b_{i+1})$ via \eqref{eq:recursive_simp_bounds} we have,
\begin{equation}\label{eq:bellman_bounds_theo}
	\begin{split}
		&\mathcal{UB}(b_i, \{a,\pi^{\star}_{(i+1)+}\}) \!=\! \mathbf{ub}(b^s_i, b_i, a) \!+ \!\!\! \expt{z_{i+1}}{\mathcal{UB}^{\star}(b_{i+1})}\\
		&\mathcal{LB}(b_{i}, \{a,\pi^{\star}_{(i+1)+}\}) \!=\! \mathbf{lb}(b^s_i, b_i, a) \!+ \!\!\!  \expt{z_{i+1}}{\mathcal{LB}^{\star}(b_{i+1})}
	\end{split}
\end{equation}
In practice, the expectation over observations is approximated by a parametric number of samples, $n_z$, which leads to 
\small
\begin{equation}\label{eq:bellman_bounds}
\begin{split}
	&\mathcal{UB}(b_i, \{a,\pi^{\star}_{(i+1)+}\}) \!=\! \mathbf{ub}(b^s_i, b_i, a) \!+ \! \frac{1}{n_z}\sum_l \mathcal{UB}^{\star}(b^l_{i+1})\\
	&\mathcal{LB}(b_{i}, \{a,\pi^{\star}_{(i+1)+}\}) \!=\! \mathbf{lb}(b^s_i, b_i, a) \!+ \!  \frac{1}{n_z} \sum_l\mathcal{LB}^{\star}(b^l_{i+1})	
\end{split}
\end{equation}
\normalsize
%
%
%
%
where 
superscript $l$ is the belief node index corresponding to the $z^l$ observation. 
The above is defined for each $a \in \mathcal{A}$.

\begin{figure}[t]
	\centering
	\includegraphics[width=\columnwidth]{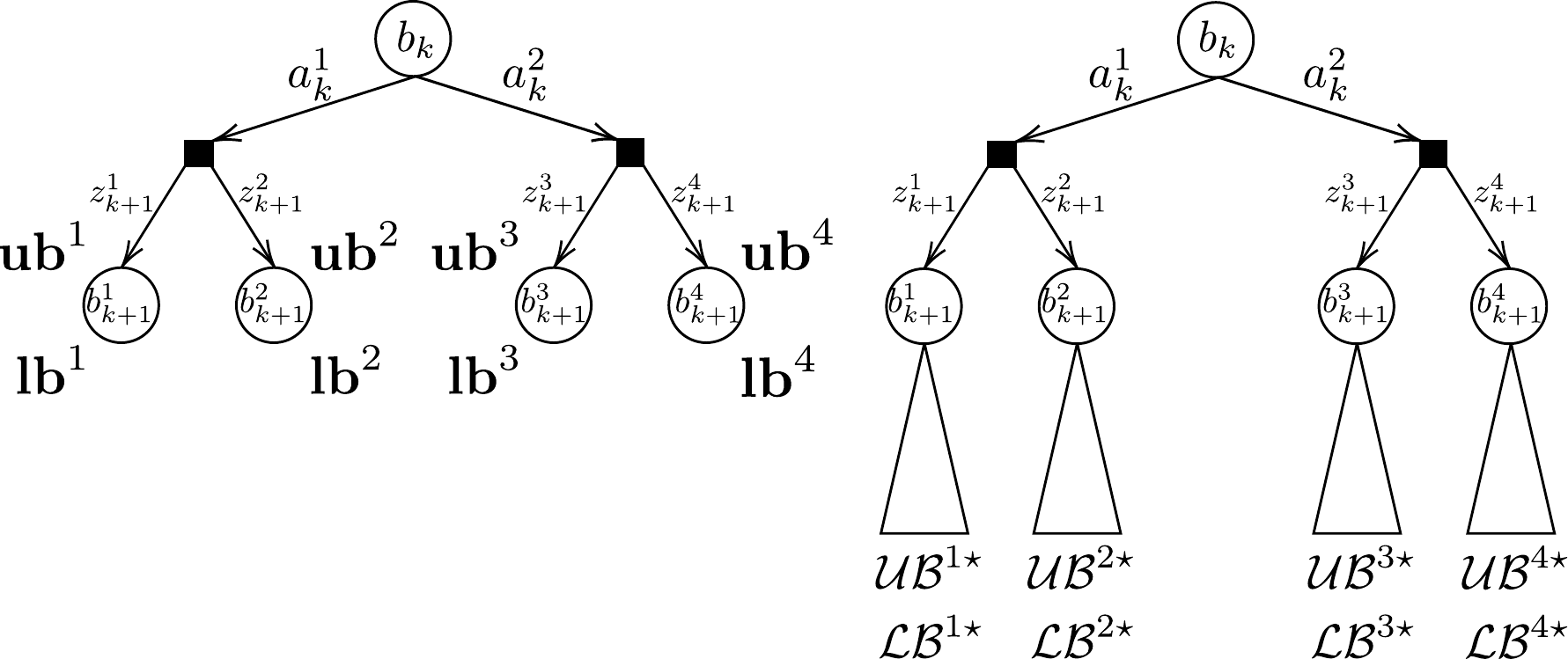}
	\caption{Left: Leaf nodes are bounded using \eqref{eq:bound_reward}. Right: Subtrees are bounded using \eqref{eq:bellman_bounds} and Alg.~\ref{alg:prune}.}
	\label{fig:bellman_bounds}
\end{figure}

Second, we perform branch pruning using Alg.~\ref{alg:prune} (see the Appendix) and as explained in Section \ref{branch-and-bound}: 
%
For each action $a\in \mathcal{A}$ we have corresponding bounds acquired via \eqref{eq:bellman_bounds}. For sufficiently tight bounds, all branches but one can be pruned 
(w.l.o.g.~the branch corresponding to action $a^{\star} \in \mathcal{A}$). Thus, the bounds corresponding to action $a^{\star}$ hold: $\mathcal{UB}^{\star}(b_i) = \mathcal{UB}(b_i, \{a^{\star},\pi^{\star}_{(i+1)+}\}), \mathcal{LB}^{\star}(b_i) = \mathcal{LB}(b_{i}, \{a^{\star},\pi^{\star}_{(i+1)+}\})$, and $\pi^{\star}_{i+}(b_i) = \{a^{\star}, \pi^{\star}_{(i+1)+}\}$. As a consequence, we get upper and lower bounds on the \emph{optimal} objective (value) function $J^{\star}(b_i)$, 
\begin{equation}\label{eq:optimal_bounds_objective}
	\mathcal{LB}^{\star}(b_{i}) \leq J^{\star}(b_i) \leq \mathcal{UB}^{\star}(b_{i}),
\end{equation}
as well as access to the optimal policy $\pi^{\star}_{i+}(b_i)$. See illustration in Fig.~\ref{fig:bellman_bounds}.

Yet, this formulation presents a difficulty. It is generally not guaranteed that after using Alg.~\ref{alg:prune} we are left with a single branch in each belief node, since the bounds might overlap (see illustration in Fig.~\ref{fig:action_graph_s0}). We discuss how we overcome this difficulty in the next section.

\subsection{Adaptive Simplification}\label{sec:Adaptive}
Simplification as defined so far is not enough. We might benefit from simplifying problem $M_A$ into $M_A^s$ whereas simplifying problem $M_B$ into $M_B^s$ might not help us. How can this happen? If for problem $M_A^s$ the bounds in \eqref{eq:bound_reward} are tight, as opposed to problem $M_B^s$ where they are loose. Thus branch pruning via Alg.~\ref{alg:prune} can preform well on $M_A^s$, but not so good on $M_B^s$. We address this difficulty by extending the definition of simplification as we envision it to be an \emph{adaptive paradigm}.


We denote  \textit{level of simplification} as how 'aggressive' the suggested simplification is. To illustrate this notion consider the belief is represented by a set of samples, as we do in Section \ref{simplification}. A \emph{coarse} simplification can correspond to accounting for only a very small subset of these particles, i.e. take a small fraction of the original set to represent the belief. A \emph{fine} simplification can correspond to a large subset of these particles, i.e.~take the majority of the samples to represent the belief. Of course we may define many levels of simplifications between the coarsest and finest levels.

We denote subscript $i$ for $s_i$ as the simplification level, where $s_0$ and $s_n$ correspond, respectively, to coarsest and finest simplification levels. Additionally we denote superscript $j$ for $s^j$ as the index corresponding to the belief's tree index. E.g. in Fig.~\ref{fig:inplace_simp} for tree node $b_{k+1}^4$ the corresponding simplification index is $s^4$ and it may assume any value of simplification: $s^4\in \{s_0, s_1, ...s_n\}$.

%
Further, we assume bounds monotonically become tighter as the simplification level is increased. More formally denote
\begin{equation}
\begin{split}\label{eq:bounds_delta}
&\overline{\Delta}^s(b,a) \triangleq \mathbf{ub}(b^s_i, b_i, a) - r(b_i,a)\\
&\underline{\Delta}^s(b,a) \triangleq r(b_i,a) - \mathbf{lb}(b^s_i, b_i, a).
\end{split}
\end{equation}

\begin{assumption}\label{eq:monotonic}
	$\forall s \in [0,n-1]$ we get:
	\begin{equation*}
	\overline{\Delta}^s(b,a) \geq \overline{\Delta}^{s+1}(b,a),\ \  \underline{\Delta}^{s}(b,a) \geq \underline{\Delta}^{s+1}(b,a).
	\end{equation*} 
\end{assumption}

Moreover, we assume the bounds for the finest simplification level $s_n$ converge to the original problem:
\begin{assumption}\label{eq:convergence}
	$\forall b_i,a$ we get:
	\begin{equation*}
		\mathbf{ub}(r^{s_n}(b_i,a)) = \mathbf{lb}(r^{s_n}(b_i,a))=r(b_i,a).
	\end{equation*}
	
\end{assumption}

A key question  is how can we decide the appropriate level of simplification beforehand? We would like the coarsest level $s_i$ that will enable eliminating actions/branches, i.e.~lead us to Fig.~\ref{fig:action_graph_s1} and not Fig.~\ref{fig:action_graph_s0}. 
%
%
If such level of simplification cannot be predetermined, then an incremental search must be carried out, i.e., try to solve the problem using the coarsest level possible and if that is not good enough move on to the next, finer level.

In Alg.~\ref{alg:simp_planning} (see the Appendix) our adaptive simplification approach is summarized. The general idea is to break down recursively a given belief tree $\mathbb{T}$ into its sub-problems (subtrees), denoted as $\{\mathbb{T}_m\}_{m=1}^{|\mathcal{A}|}$, and solve each sub-problem with its own simplification level $s_i$. Ultimately this would lead to the solution of the entire problem via \eqref{eq:bellman_bounds}. 

A potential computational issue with this formulation is that numerous repeating calculations for several levels of simplifications might not worth it, since the overall time for all levels calculations is suppressing the time it takes to solve the original problem. Fortunately, this is not an issue with our adaptive simplification, as discussed next.


Our adaptive simplification approach is based on two key observations. 
The \emph{first key observation} is that we can compare bounds from \emph{different} levels of simplification when pruning. Our \emph{second key observation} is that we can re-use calculations between different simplification levels, and thus avoid re-calculating simplification from scratch.  In the following sections we elaborate on each of these crucial aspects.

\subsubsection{Comparing Bounds with Different Simplification Levels}\label{sec:ComBoundsAdapt}

\begin{figure}[t]
	\centering
	\includegraphics[width=0.4\columnwidth]{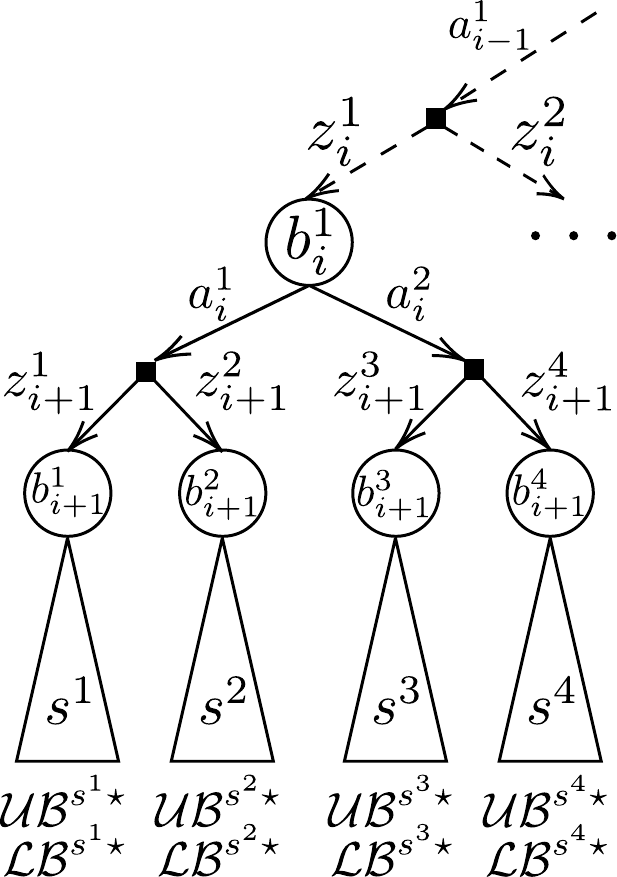}
	\caption{Adaptive Simplification}
	\label{fig:adaptive_bounds}
\end{figure}

Consider again some belief node $b_i$ in the belief tree, and assume recursively for \emph{each} of its children belief nodes $b_{i+1}$ we already calculated  the optimal policy $\pi_{(i+1)+}^{\star}(b_{i+1})$ and the corresponding upper and lower bounds $\mathcal{UB}^{s^l \star}(b_{i+1})$ and $\mathcal{LB}^{s^l \star}(b_{i+1})$, where $s^l$ indicates the simplification level, and $l$ corresponds to the belief tree nodes indexing notation. In general, the bounds for each belief node $b_{i+1}$ can correspond to different simplification levels, as illustrated in Fig.~\ref{fig:adaptive_bounds}. 

We now discuss how the simplification level is updated recursively, and revisit the process to calculate the optimal policy and the corresponding bounds for belief node $b_i$, previously described by Eqs.~\eqref{eq:bellman_bounds} and \eqref{eq:optimal_bounds_objective}. Incorporating adaptive simplification, Eq.~\eqref{eq:bellman_bounds} is modified to 
\begin{equation}\label{eq:bellman_bounds_simp_level}
	\begin{split}
		&\mathcal{UB}^{s^j}(b_i, \{a^j_i,\pi^{\star}_{(i+1)+}\}) \!=\! \mathbf{ub}(b^s_i, b_i, a^j_i) \!+ \!\! \frac{1}{n_z}\sum_l\mathcal{UB}^{s^l \star}(b^l_{i+1})\\
		&\mathcal{LB}^{s^j}(b_{i}, \{a^j_i,\pi^{\star}_{(i+1)+}\}) \!=\! \mathbf{lb}(b^s_i, b_i, a^j_i) \!+ \!\! \frac{1}{n_z}\sum_l \mathcal{LB}^{s^l \star}(b^l_{i+1}).
	\end{split}
\end{equation}
Note this equation applies for each $a^j_i\in \mathcal{A}$, and as mentioned, each belief node $b^l_{i+1}$ (one for each observation $z^l_{i+1}$) has its own simplification level $s^{l}$.  In other words, for  each  $b^l_{i+1}$, $s^{l}$ is the simplification level that was sufficient for calculating the bounds $\{\mathcal{UB}^{s^l \star}(b^l_{i+1}), \mathcal{LB}^{s^l \star}(b^l_{i+1})\}$  and the corresponding optimal policy $\pi_{(i+1)+}^{\star}(b^l_{i+1})$. Thus, when addressing belief node $b_i$ in \eqref{eq:bellman_bounds_simp_level}, for each belief node $b^l_{i+1}$ and its corresponding simplification level $s^l$, these bounds  are already available.  Yet, we may still need to adapt the simplification level as we further discuss in Section \ref{seq:adaptive_reuse}.

Further, as seen in \eqref{eq:bellman_bounds_simp_level}, the immediate reward and the corresponding bounds $\mathbf{ub}$ and $\mathbf{lb}$, in general, can be calculated with their own simplification level $s$. In particular, when starting the calculations, $s$ could correspond to some default coarse simplification level, e.g.~coarsest simplification level $s_0$.

To define simplification level $s^j$ of the bounds  \eqref{eq:bellman_bounds_simp_level} we remind the belief tree is a discrete approximation to the expectation taken w.r.t future observations $z_i, i \in \{k,k+L\}$. We account for some number $n_z$ of observations made in tree nodes (e.g. in Fig.~\ref{fig:adaptive_bounds}, $n_z=2$)
\begin{equation}\label{eq:sj}
	s^j\triangleq \min\{s, s^{l_1}, s^{l_2},...s^{l_{n_z}}\},
\end{equation}
where $\{s^{l_1}, s^{l_2},...s^{l_{n_z}}\}$ represents the (generally different) simplification levels of belief nodes $b^l_{i+1}$ considered in the expectation approximation in \eqref{eq:bellman_bounds_simp_level}. We explain the reason to define $s^j$ as such in Section~\ref{seq:adaptive_reuse}.

As earlier, we wish to decide which action $a_i^{\star}\in\mathcal{A}$ is optimal from belief node $b_i$; the corresponding optimal policy would then be $\pi_{i+}^{\star} = \{a_i^{\star}, \pi^{\star}_{(i+1)+}\}$, where $\pi^{\star}_{(i+1)+}$ is the already-calculated optimal policy for belief node $b^l_{i+1}$ that $a_i^{\star}$ leads to.  See illustration in Fig.~\ref{fig:adaptive_bounds}. 

Determining $a_i^{\star}$ requires eliminating all other candidate actions $a^j \in \mathcal{A}$, which involves comparing their  corresponding bounds \eqref{eq:bellman_bounds_simp_level}. Importantly, the bounds are analytical, i.e.~they are \emph{valid for all simplification levels}. 

As earlier, we can compare bounds for different candidate actions and if the bounds do not overlap, perform pruning. For example, if for $a^1_i,a^2_i \in \mathcal{A}$, 
\begin{equation}\label{eq:example_prun2}
	\mathcal{UB}^{s^{1}}(b_i, \{a^1_i,\pi^{\star}_{(i+1)+}\})  < \mathcal{LB}^{s^{2}}(b_i, \{a^2_i,\pi^{\star}_{(i+1)+}\}), 
\end{equation}
we can prune the $a_i^1$ branch. Note how pruning was carried out while the sibling subtrees have completely different levels of simplification.

If the bounds are sufficiently tight and all branches but one were pruned, then the remaining action, in this case $a_i^2$, is announced as $a_i^{\star}$, and $s^2$ is announced as $s^{\star}$. Thus the above-mentioned optimal policy $\pi^{\star}_{i+}$ is constructed. 

We now recall that $b_i$ by itself has an index in the belief tree, with respect to the previous level.  We thus denote it as $b_i^l$, considering the father node of $b_i$ is $b_{i-1}$,  and the $l$th observation $z^l_i$.

At this point we get  
\begin{equation}\label{eq:bellman_bounds_simp_level_optimal}
\begin{split}
&\mathcal{UB}^{s^l \star}(b^l_i) = \mathcal{UB}^{s^{\star}}(b_i^l, \{a^{\star},\pi^{\star}_{(i+1)+}\})\\
&\mathcal{LB}^{s^l \star}(b^l_i) = \mathcal{LB}^{s^{\star}}(b_{i}^l, \{a^{\star},\pi^{\star}_{(i+1)+}\}).
\end{split}
\end{equation}
%
Similarly to \eqref{eq:optimal_bounds_objective}, \eqref{eq:bellman_bounds_simp_level_optimal} leads to bounds over the objective function

\begin{equation}
	\mathcal{LB}^{s^l \star}(b^l_i) \leq J^{\star}(b^l_i) \leq \mathcal{UB}^{s^l \star}(b^l_i),
\end{equation}
where $b^l_i$ corresponds to the same notation as in \eqref{eq:bellman_bounds_simp_level}, recursively. Crucially, these bounds are valid for \emph{any} simplification level $s^l \in \{s_0,\ldots,s_n\}$; thus, we can compare between subtrees with different simplification levels.


In general, pruning as done in \eqref{eq:example_prun2} will not always hold, and we need to adapt level of simplification. We discuss in the next section how we do this, including re-use of calculations.

\subsubsection{Adapting Simplification Level with Calculation Re-Use}\label{seq:adaptive_reuse}
For some belief node $b_i$ in the belief tree, consider the bounds 
 $\mathcal{UB}^{s^j}(b_i, \{a^j_i,\pi^{\star}_{(i+1)+}\})$ and $\mathcal{LB}^{s^j}(b_{i}, \{a^j_i,\pi^{\star}_{(i+1)+}\})$ from \eqref{eq:bellman_bounds_simp_level} for different actions $a^j_i \in \mathcal{A}$,  that partially overlap and therefore could not be pruned. Each such action $a^j_i$ can generally have its own simplification level $s^j$. 
 
 We now iteratively increase the simplification level by $1$. This can be done for each of the branches, if $s^j$ is identical for all branches, or only for the branch with the coarsest simplification level.
 
 Consider now any such branch whose simplification level needs to be adapted from $s^j$ to $s^j+1$. 
  Recall, that at this point, the mentioned bounds were already calculated, thus their ingredients, in terms of 
 $\mathbf{ub}(r^s(b_i,a^j))$, $\mathbf{lb}(r^s(b_{i},a^j))$ and  $\{ \mathcal{UB}^{s^l \star}(b_{i+1}) , \mathcal{LB}^{s^l \star}(b_{i+1})\}_{l=1}^{n_z}$, involved in approximating the expectation in  \eqref{eq:bellman_bounds_simp_level}, are available. Recall also $s^j\triangleq \min\{s, s^{l_1}, s^{l_2},...s^{l_{n_z}}\}$ from \eqref{eq:sj}, i.e.~each element  in $\{s, s^{l_1}, s^{l_2},...s^{l_{n_z}}\}$ is either equal or larger than $s^j$. We now discuss both cases, starting from the latter.
 
As we assumed bounds to improve monotonically as simplification level increases, see Assump.~\ref{eq:monotonic}, for any $s^l>s^j+1$ we already have readily available bounds $\{ \mathcal{UB}^{s^l \star}(b_{i+1}) , \mathcal{LB}^{s^l \star}(b_{i+1})\}$
which are tighter than those that would be obtained for simplification level $s^j+1$. Thus, we can \emph{safely skip} the calculation of the latter and use the existing bounds from level $s^l$ as is. The same argument also applies for bounds over momentary rewards, i.e.~$\mathbf{ub}(r^s(b_i,a^j))$ and $\mathbf{lb}(r^s(b_{i},a^j))$.

For the former case, i.e.~$s^l = s^j$, we now have to adapt the  simplification level to $s^j+1$ by calculating the bounds $\{ \mathcal{UB}^{(s^l+1) \star}(b_{i+1}) , \mathcal{LB}^{(s^l+1) \star}(b_{i+1})\}$. Here, our \emph{key insight} is that, instead of calculating these bounds from scratch, we can re-use calculations between different simplification levels, in this case, from level $s^l$. As the bounds from that level are available, we can identify only the incremental part that is "missing" to get from simplification level $s^l$ to $s^l+1$, and update \emph{analytically} the existing bounds $\{ \mathcal{UB}^{s^l \star}(b_{i+1}),  \mathcal{LB}^{s^l \star}(b_{i+1})\}$ to recover $\{ \mathcal{UB}^{(s^l+1) \star}(b_{i+1}),  \mathcal{LB}^{(s^l+1) \star}(b_{i+1})\}$ exactly. The same argument applies also for bounds over momentary rewards. In section \ref{reuse_calculations_entropy} we apply this approach to a specific simplification and reward function.

We can repeat iteratively the above process of increasing the simplification level until we are able to prune all branches but one. Note this means each subtree will be solved maximum once, per simplification level. Since we assumed the simplification converges to the original problem for the finest level $s_n$, see \eqref{eq:convergence}, we are guaranteed to eventually disqualify all sub-optimal branches. Moreover, due to the discussed-above calculation re-use, in the worst case, adapting the simplification all the way up to the finest level $s_n$, is roughly equivalent to solving the original problem. We address this aspect explicitly in Section \ref{reuse_calculations_entropy}. For a detailed illustrative example of procedure described, accounting for Fig.~\ref{fig:adaptive_bounds} see  Appendix~\ref{appen:adaptive_simulation}.

In the following sections we present a specific simplification along with novel derivations that show it holds the mentioned mathematical properties.

\subsection{Bounds}\label{sec:Bounds}
We now dive deeper to the bounds our chosen simplification suggests. In our setting we simplify each belief $b$ into $b^s$. There are some ways to do so, e.g. for a Gaussian case we can sparsify the covariance or information matrix, while in a  sampling-based belief representation we can choose a subset of samples. The trick is to mathematically bound the reward function using the simplified and possibly the original beliefs, as in \eqref{eq:bound_reward_general}.
In the following subsections we derive such bounds.

\subsubsection{Lipschitz Continuous Rewards}\label{LC_bounds_section}
Given a Lipschitz continuous reward function (see definition in Appendix~\ref{appen:LC_definition}), denoted as LC, we can upper and lower bound it like so:
\begin{equation}\label{eq:LC_rewards}
r(b^s,a) - \lambda_r d(b,b^s) \leq r(b,a) \leq r(b^s,a) + \lambda_r d(b,b^s),
\end{equation}
where $d(b,b^s)$ is a distance metric between two posterior distributions (e.g. symmetric KL divergence, EMD, $L_1$ etc.). This formulation suggests we, using the LC properties of the reward function, can now upper and lower bound the true objective function approximated by the belief tree. This is also true for any belief node in any depth of the belief tree.
\begin{theorem}\label{theorem:LC_bounds}
	Given an LC reward function and a simplification operator \eqref{eq:reward_simplifcation}, the objective in any depth can be upper and lower bounded as
	\begin{equation}
		\mathcal{LB}(b_i, \pi_{i+}) \leq J(b_i,\pi_{i+}) \leq
		\mathcal{UB}(b_i, \pi_{i+}).
	\end{equation}
\end{theorem}
We prove this theorem and provide explicit expressions for the upper and lower bounds in Appendix \ref{proof:theorem_1}.

Further, using \eqref{eq:bellman_bounds} we get a recursive update for LC bounds by setting:
$\mathbf{lb} = r(b^s,a)- \lambda_r \cdot d(b, b^s)$ and $\mathbf{ub} = r(b^s,a)+ \lambda_r \cdot d(b, b^s)$.

%
This formulation is true for any LC reward function. Although it can be expanded to more general reward functions such as the $\alpha$-H\"{o}lder. in this paper we focus on information theoretic rewards. However, unfortunately, most prominent of those such as the differential entropy are \emph{not} LC or $\alpha$-H\"{o}lder. Next, we derive novel \emph{analytical} bounds for differential entropy, considering a sampling based belief representation.
\subsubsection{Belief Representation and Chosen Simplification}\label{simplification}
As mentioned earlier, we  use Particle Filter for planning. This means the belief is represented as a set of weighted particles, 
\begin{equation}\label{eq:belief_particles}
b \triangleq \{x^i, w^i\}_{i=1}^N,
\end{equation}
where $N$ is a tune-able parameter specifying the desired number of particles. Maintaining the weights is done via Importance Sampling. 

\emph{Suggested Simplification:} Given the belief representation  \eqref{eq:belief_particles}, the simplified belief is some subset of $N^s$ particles, sampled from the original belief, where $N^s \leq N$. More formally:
\begin{equation}\label{eq:simplification}
	b^s \triangleq \!\!\{(x^i, w^i)\mid i \in A^s, A^s\subseteq \{1,2,\ldots,N\}, \! |A^s|=\! N^s \},
\end{equation}
where $A^s$ is the set of particle indices comprising the simplified belief $b^s$. Further, we denote $A_k^s$ for this set corresponding to the simplified belief $b_k^s$ from time instant $k$ .

Increasing the level of simplification is done \emph{incrementally}. Specifically, consider $|A^{s}|=N^{s}$ and to get to the next simplification level we add $m$ particles with indices $j \in B, \ |B|=m$. Then the following holds:  $A^{s} \cap B=\emptyset, \ A^{s+1}= A^{s} \cup B, \  N^{s+1}=N^{s}+m$.

\subsubsection{Bounding the Differential Entropy}\label{bounding_entropy_section}
As mentioned, for the non-LC case a less general approach must be taken. Here we consider a common reward function, the differential entropy, which indeed is not LC. As one of our key contributions, in this section we derive novel analytical upper and lower bounds $\mathbf{lb}$ and $\mathbf{ub}$ considering this reward function, assuming a sampling-based belief representation \eqref{eq:belief_particles} and the corresponding simplified belief \eqref{eq:simplification}. These bounds can then be used within our general simplification framework presented in Sections \ref{sec:simplification} and \ref{sec:Adaptive}.

Under this setting approximating the differential entropy of the belief is not an easy task. To calculate $\mathcal{H}(b[x_k])=-\int b[x_k]\cdot \log\left(b[x_k]\right) dx_k$, one must have access to the manifold representing the belief. Several approaches exist. One of them is using Kernel Density Estimation as done, e.g., by \citet{Fischer20icml} which costs $O(d\cdot N^2)$, where $d$ is the dimension of the problem. In this paper we consider the method proposed by \Citet{Boers10fusion}:
\begin{flalign}\label{eq:boers_diff_ent}
&\hat{\mathcal{H}}(b_{k+1}) \triangleq \ubrace{\log \left[\sum_i \prob{z_{k+1}\mid x_{k+1}^i} w_k^i\right]}{(a)}\\
&\ubrace{-\sum_i w_{k+1}^i\cdot\log\left[\prob{z_{k+1}\mid x_{k+1}^i} \sum_j\prob{x_{k+1}^i\mid x_k^j, a_k}w_k^j\right]}{(b)},\notag
\end{flalign}
where $i$ indexes particles and their corresponding weight as in \eqref{eq:belief_particles} and \eqref{eq:simplification}.
One can observe this method requires access to samples representing both $b_k$ and $b_{k+1}$; thus, this corresponds to an information-theoretic reward of the form $r(b_k,b_{k+1})$. 

%
Utilizing the chosen simplification \eqref{eq:simplification} we can now upper and lower bound \ref{eq:boers_diff_ent}. We do so below by bounding separately the  terms (a) and (b) it is comprised from.


First, as the models are known, we define 
$m \triangleq \max \{\prob{x_{k+1} \mid x_k, a_k}\}$  and  $n \triangleq \max \{\prob{z_{k}\mid x_{k}}\}$.

\begin{theorem}\label{theorem:lower_upper_I}
	Term $(a) $ in \eqref{eq:boers_diff_ent} can be upper and lower bounded via simplification as,
	\small
	\begin{equation*}
	\begin{split}
	&(a) \geq
	\log \left[\sum_{i\in A^s_{k+1}} \prob{z_{k+1}\mid x_{k+1}^i} w_k^i \right]\\
	&(a) \! \leq \!
	\log \left\{\!\! \sum_{i\in A^s_{k+1}} \prob{z_{k+1}\mid x_{k+1}^i} w_k^i + \!
		n\cdot \left[1-\sum_{i\in A^s_{k+1}}w_{k}^i\right] \!\! \right\}
	\end{split}
	\end{equation*}
	\normalsize
\end{theorem}

See proof in  Appendix \ref{proof:theorem_2}.

\begin{theorem}\label{theorem:lower_upper_II}
	Term $(b)$ in  \eqref{eq:boers_diff_ent} can be upper and lower bounded via simplification as
	\small
	\begin{equation*}
	\begin{split}
	&(b)\geq  -\sum_{i \in \neg A^s_{k+1}} w_{k+1}^i\cdot\log\left[ m \cdot \prob{z_{k+1}\mid x_{k+1}^i}\right]\\
	&-\sum_{i \in A^s_{k+1}} w_{k+1}^i\cdot\log\left[\prob{z_{k+1}\mid x_{k+1}^i} \sum_j\prob{x_{k+1}^i\mid x_k^j, a_k}w_k^j\right]\\
	&(b) \leq
	-\sum_i w_{k+1}^i\cdot\log\left[\sum_{j \in A^s_k} \prob{z_{k+1}\mid x_{k+1}^i} \prob{x_{k+1}^i \mid x_k^j, a_k}w_k^j\right] 
	\end{split}
	\end{equation*}
	\normalsize
\end{theorem}
See proof in  Appendix \ref{proof:theorem_3}.

Finally,  bounding \eqref{eq:boers_diff_ent} using Theorems  \ref{theorem:lower_upper_I} and \ref{theorem:lower_upper_II} corresponds, in our general framework from Sections \ref{sec:simplification} and \ref{sec:Adaptive}, to  \eqref{eq:bounds_extended}: $\mathbf{lb}(b_k^s, b_{k+1}^s, b_k,b_{k+1}) \leq	r(b_k, b_{k+1}) \leq  \mathbf{ub}(b_k^s, b_{k+1}^s, b_k,b_{k+1})$. 
%
%
%

\subsection{Bounds Analysis}

\subsubsection{Convergence}

We now analyze the convergence of the simplification described  in Section \ref{simplification}. Since simplifying to the level of $s_n$ means the simplified belief is just the original belief, we get:
$N^s=N \Rightarrow b=b^s \Rightarrow r(b^s,a) = r(b,a), d(b, b^s)=0$
Furthermore, it can be easily seen the upper and lower bounds in Theorem \ref{theorem:lower_upper_I}  coincide and equal to term (a) from \eqref{eq:boers_diff_ent}, and, similarly,  the upper and lower bounds in Theorem \ref{theorem:lower_upper_II} coincide and equal to term (b) from \eqref{eq:boers_diff_ent}. Meaning, under our chosen simplification the bounds converge to the original rewards for the non simplified original belief. 
\subsubsection{Complexity Analysis}
The original formulation of \citet{Boers10fusion} suggests complexity of $O(N^2)$ where $N$ is as in \eqref{eq:belief_particles}. The derivations in Section \ref{bounding_entropy_section} suggest complexity of $O(N^s \cdot N)$.

On the other hand, analyzing time complexity of the derivations in Section  \ref{LC_bounds_section} is not straightforward. We would like to have $O(r(b^s, a) + d(b, b^s)) < O(r(b,a))$. Although using $N^s$ particle surely results $O(r(b^s, a)) < O(r(b,a))$ the exact complexity reduction depends directly on the specific reward function used. We also must account for the distance term $d(b, b_s)$. In order for us to achieve speedup we must have $O(d(b, b^s)) < O(r(b,a))$. So, when using certain reward functions, the chosen metric must be sub-complex to the reward function. Fortunately such metrics exist, e.g.~$L_1$ which is $O(N)$ complex. 

Altogether, our suggested approach may speedup performance from $O(N\cdot N)$ to $O(N\cdot N^s)$. Since $N^s$ is the parametric number of particles, we can control the desired speedup. E.g if $N=100$ we could set $N^s=10$. In such setting, if re-simplification takes no place, potential speedup is by a factor of $\times 10$. If re-simplification is needed, the speedup will be smaller, but in any case, without compromising on the solution's quality. 

\subsubsection{Re-use of Calculations}\label{reuse_calculations_entropy}
%

The unique structure of the simplification allows us to cache the calculation from a previous simplification level and add or subtract what we need in order to get the terms for the current simplification level. Specifically, consider moving from  simplification level $s$ to  level $s+1$, and assume this corresponds to adding $m$ additional particles, i.e.~$N^{s+1}=N^{s}+m$. In other words,  
%
the new simplified belief $b^{s+1}$ should have additional $m$ particles compared to $b^{s}$. 


 For the terms in Theorem \ref{theorem:lower_upper_I} and lower bound in Theorem \ref{theorem:lower_upper_II} we can just cache the final result and augment the sums with the calculations corresponding to the $m$ missing particles. The upper term in Theorem \ref{theorem:lower_upper_II} is more complex. We need to cache the inner sums of the $\log$, augment them with the missing $m$ particles calculations and re-weight it using the outer sum weights. Although less straightforward, this is still very simple to do and space complexity remains the same since we are already caching the particles and weights for each belief in the corresponding belief node.
 
%
%
%

This re-use of calculations results in time complexity, going up from simplification level $s_0$ to level $s_n$ being the same order as just solving the original problem in the first place. Thus making simplification is worthwhile always (up to some constant overhead). Moreover, it is sufficient that at least for some of the belief nodes $b_i$ in the belief tree we are able to identify optimal policies without reaching simplification level $s_n$ for our approach to exhibit a computational speedup. Of course, coarser simplification levels  correspond to more significant speedup. We demonstrate these aspects explicitly in Section \ref{experiment_planning}. 


\section{Experimental Setting and Results}
In this section we examine two experimental settings that show: (a) How appealing and attractive our novel differential entropy bounds can be. (b) How they can use us to speed up POMDP planning as explained in Section \ref{approach}.
All experiments were conducted on a laptop with Intel i7-9850H CPU 2.59GHz with 16 GB RAM and no parallelism.
\subsection{Differential Entropy Approximations}\label{experiment_entropy}
We design a simple experiment that shows how our novel differential entropy bounds from Section \ref{bounding_entropy_section} can exhibit a tight behavior, considering a setting of an agent moving in a known 2D environment represented by spatially scattered beacons. 
	The agent has some prior belief $b[x_0]$ over its initial position $x_0\in \mathbb{R}^2$. In this experiment the agent moves according to a predefined policy, i.e.~in this experiment planning is not involved. 	
	At each time step the agent executes the action determined by the given policy and acquires a relative distance observation $z\in \mathbb{R}^2$ to the closest beacon. The closer the agent is getting to a beacon the less 'noisy' the measurement is going to be, i.e. getting closer to a beacon reduces the agent's uncertainty in the world. We consider a sampling-based belief representation and use a standard particle filter with importance sampling to recursively update the belief. 	The belief at each time step is thus represented by a set of $N$ weighted samples as in \eqref{eq:belief_particles}. 

\begin{figure}[!ht]
	\centering
	\subfloat[ \label{fig:entropy_simulation}]{\includegraphics[width=0.45\columnwidth]{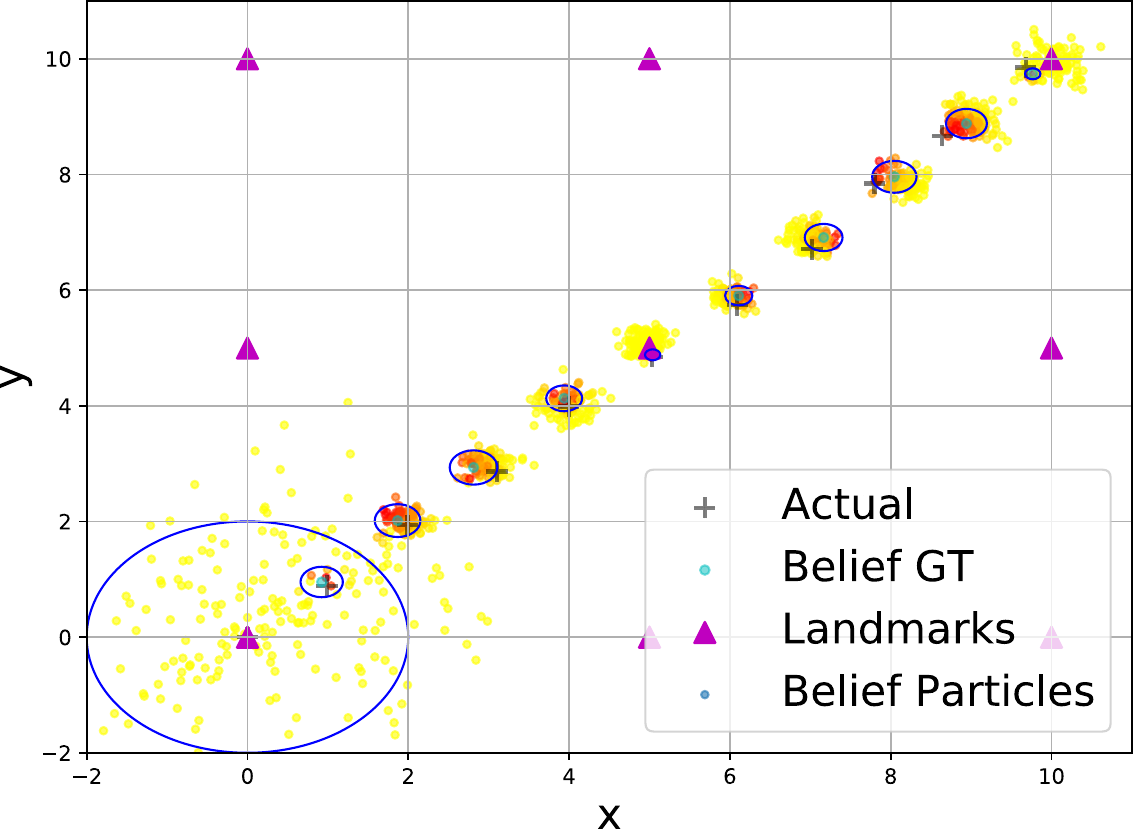}}
		\qquad
	\subfloat[ \label{fig:plannig_simulation}]{\includegraphics[width=0.45\columnwidth]{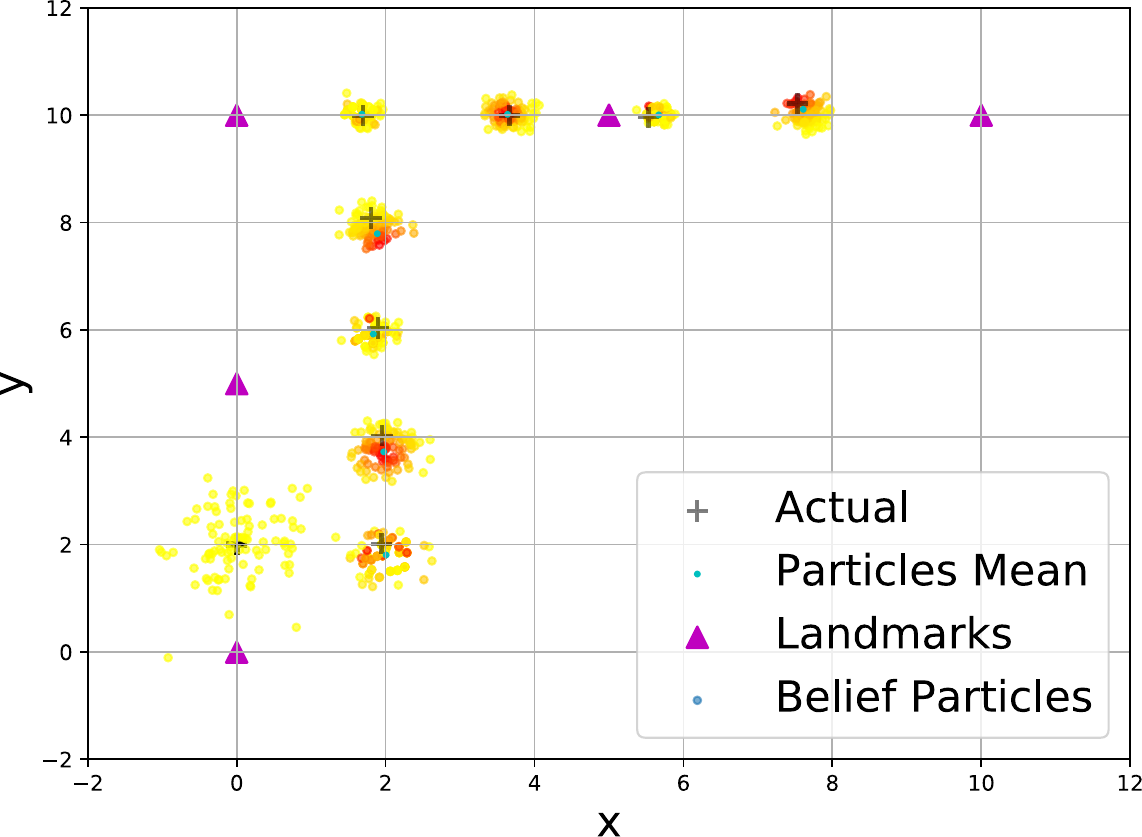}}
	\caption{\textbf{(a)} Localization with a predefined policy in an environment represented by known landmarks. Particles are colored according to weight (high-red, small-yellow). Blue ellipses correspond to estimated uncertainty covariance by a KF. \textbf{(b)} Planning simulation, setting II. At each time step the agent is planning $L$ step into the future and then executes the first action from the calculated optimal policy. The plot shows the belief evolution in terms of particles sets along the actual trajectory taken by the agent.}
\end{figure}

\begin{figure*}[th]
	\centering
	\includegraphics[width=1.6\columnwidth]{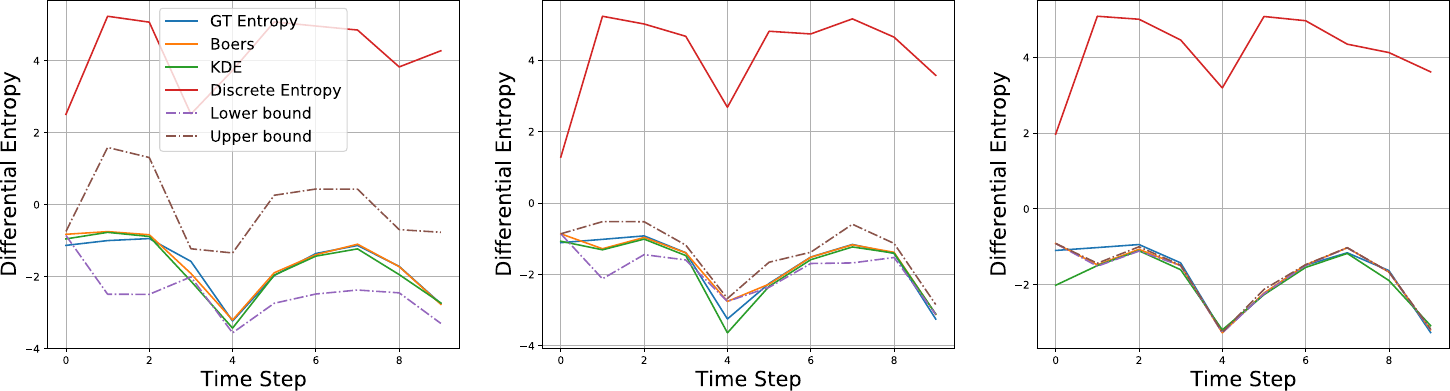}
	\caption{Differential entropy approximations and bounds.  Calculations were done using $N=200$ particles. From left to right: Simplification is $N^s=\{0.1, 0.5, 0.9\}\cdot N$.}
	\label{fig:ent_bounds}
\end{figure*}
In this study we examine the bounds convergence to the  entropy approximation \eqref{eq:boers_diff_ent}. We also assert the approximation itself is: (1) Satisfying compared to the real analytical differential entropy, and (2) competitive w.r.t. other recently used methods i.e. Kernel Density Estimation (KDE).

	To achieve the desired comparison we need to account for the real analytical differential entropy, entropy approximation according to KDE, entropy approximation according to  \eqref{eq:boers_diff_ent}, and our suggested bounds. Additionally, we also show the naive approximation using discrete entropy of the particles weights: $h(b)=-\sum_i w^i \cdot \log w^i$.
	
	To show the above we set the transition and observation models, $T$ and $\mathcal{O}$, to be linear with additive white Gaussian noise: $T=\prob{x' \mid x, a}=\mathcal{N}(x+a, I\cdot \sigma_T)$, $\mathcal{O}=\prob{z \mid x}=\mathcal{N}(x-x^b, I\cdot\sigma_{\mathcal{O}}\cdot \max\{r,r_{min}\})$, where  $r$ is the agent's distance to the nearest beacon whose known location is $x^b$, and $r_{min}$ is a tune-able parameter. The observation model suggests the noise level is proportional to the agent's distance from the nearest beacon.
	Further, we set the initial belief as a 2-D Gaussian $b[x_0] = \mathcal{N}(x_0, I\cdot \sigma_0)$.

	This setting implies the belief at each time step is in fact a Gaussian and it can be inferred exactly using Kalman Filter (KF). Since the Gaussian distribution has a closed form expression for differential entropy, we can calculate the real analytical differential entropy.

	For all other approaches in this study, we use a sampling based belief representation \eqref{eq:belief_particles}, and update it via a standard particle filter. Initial belief particles are sampled from $b[x_0]$ as it is defined above. 

	Thus we utilize the sampling-based belief representation  	
	to calculate the entropy approximation \eqref{eq:boers_diff_ent} and our proposed bounds \ref{bounding_entropy_section}. For the KDE approach, we use an of-the-shelf KDE package to reconstruct the belief manifold.
	

The considered scenario is shown in Fig.~\ref{fig:entropy_simulation} the agent is moving diagonally from bottom left point of the map to upper right. Along the way it  comes very near to two beacons. As a result the uncertainty is reduced accordingly. 
In Fig.~\ref{fig:ent_bounds} it can be seen how the bounds are converging towards the real approximation of the entropy. Furthermore these results suggest both KDE and \citet{Boers10fusion} approximations are doing well.
Motivated by real world problems we would like to emphasize how the approximation  scales with increasing number of particles. See the Appendix \ref{sec:supplementary} of how the bounds converge using different number of particles and how with increasing number of particles the bounds are getting tighter with coarser level of simplification. Although a dozen or two of particles is enough for this toy example of a 2D setting, moving on to real-world problems such as 3D localization which has 6-DOF will require many more particles to account for the real posterior distribution.

\setlength{\tabcolsep}{5pt}
\begin{table*}
	\scalebox{0.82}{ 
		\begin{tabular}{l c  ccc |c ccc |c ccc}                             
			\hline                                                       
			\multicolumn{1}{c}{\multirow{ 1}{*}{Simulation}} &  \multicolumn{1}{c}{\multirow{ 1}{*}{Horizon}} & \multicolumn{3}{c}{\citet{Ye17jair} Tree} & \multicolumn{1}{c}{\multirow{ 1}{*}{Horizon}} & \multicolumn{3}{c}{\citet{Lim20ijcai} Tree} &  \multicolumn{1}{c}{\multirow{ 1}{*}{Horizon}} & \multicolumn{3}{c}{\citet{Silver10nips} Tree}\\
			\hline
			{} & {} & \multicolumn{1}{c}{20} &  \multicolumn{1}{c}{50} & \multicolumn{1}{c}{100} & \multicolumn{1}{c}{} & \multicolumn{1}{c}{10} &  \multicolumn{1}{c}{20} & \multicolumn{1}{c}{30} & \multicolumn{1}{c}{} &\multicolumn{1}{c}{20} &  \multicolumn{1}{c}{50} & \multicolumn{1}{c}{100}\\
			\cline{3-13}                                                      
			\multirow{3}{*}{Setting I} 	 & 1 	&	0.124/\textbf{0.043} & 0.741/\textbf{0.192} & 2.892/\textbf{0.667} & 1 & 0.554/\textbf{0.287} & 4.065/\textbf{1.437} & 12.908/\textbf{3.953} & 5& 1.13/\textbf{0.776} & 6.625/\textbf{2.008} & 28.19/\textbf{7.232}\\ 
			{} 	 & 2	& 	 0.364/\textbf{0.129} & 2.196/\textbf{0.584} & 8.616/\textbf{2.042} &  2 & 11.02/\textbf{5.386} & - & - & 10 & 	2.648/\textbf{2.555} & 15.342/\textbf{8.214} & -\\ 
			{} 	 & 3	& 	 0.853/\textbf{0.339} & 5.059/\textbf{1.324} & 19.899/\textbf{4.658} & 3&  - & - & - &15 & 	 4.2/\textbf{3.677} & 26.205/\textbf{20.174} & -\\ 
			
			\hline
			
			\multirow{4}{*}{Setting II} 	 & 1 	&  0.245/\textbf{0.099} & 1.513/\textbf{0.4} & 5.855/\textbf{2.018} & 1&  1.112/\textbf{0.953} & 8.501/\textbf{5.143} & 26.375/\textbf{11.977}& 5 & 1.383/\textbf{0.733} & 8.417/\textbf{3.864} & 33.244/\textbf{10.97}\\ 
			{} 	 & 2 	& 	 1.209/\textbf{0.738} & 7.195/\textbf{3.821} & 30.638/\textbf{13.49} & 2 & - & - & - &10 & 2.985/\textbf{2.112} & 17.293/\textbf{6.092} & -\\  
			{} 	 & 3 	& 	 5.027/\textbf{3.212} & 31.515/\textbf{18.288} & - &3 & - & - & - & 15 & 	 4.53/\textbf{3.701} & 27.712/\textbf{11.385} & -\\

			\hline                                                 
	\end{tabular}}                                          
	\caption{Mean time per planning session. For each table entry results are (\textit{original objective time/simplified approach time}). Results are in Seconds. Second row in the table indicates the number of particles used. DESPOT-like belief tree \cite{Ye17jair} was built by expanding all actions in every level of the tree but only making a single observation in each level, i.e.~$n_z=1$. The corresponding belief tree's structure is similar to DESPOT's tree: 'thick' on action space but sparse on the observation space. POWSS-like tree \cite[Alg.~3]{Lim20ijcai} was built by expending each belief node for all actions, and 		
		generating an observation for each particle of the belief, i.e.~number of observations when branching is as the number of particles. This corresponds to a belief tree that is 
		'thick' on action and observation spaces. POMCP-like tree \cite{Silver10nips} was built using 5 'rollout's starting from the root of the tree. In each rollout down the tree we randomly choose if to expand a new node by taking an action that was not taken previously from that node or to go down the tree using nodes that were already expanded from previous rollouts. Thus we get a belief tree with structure similar to the POMCP one: several 'deep' laces, i.e. sparse on actions and observation spaces.}
	\label{table:DESPOT_Lim__POMCP_table}
\end{table*}

\subsection{Planning in 2D environment}\label{experiment_planning}
We demonstrate the gained speed up when using simplification on a 2D scenario. The world setup is the same as in Section \ref{experiment_entropy} but this time the agent is given a destination point $x^t$ and it needs to find the optimal way to get there  while reducing uncertainty. In order to do so the agent is given the ability to plan a parametric number of $L$ steps into the future. When planning session is done, the agent executes the first action out of the calculated optimal policy. After the action execution, the agent acquires a new observation, updates its belief, and re-plans again L steps into the future. This process  is repeated until the end of the scenario is reached.  Reward function is  
\begin{equation}
-r(b,a)=\expt{x \sim b}{ \| x - x^t\|_1 }+\hat{\mathcal{H}}(b),
\end{equation}
where the first term is the expected distance to goal, and $\hat{\mathcal{H}}(b_k)$ is the approximation of the differential entropy \eqref{eq:boers_diff_ent}.


To show the advantage of our approach we show it can be applied to different given trees. We evaluate our approach on a POMCP-like tree (deep and sparse) \cite{Silver10nips}, DESPOT-like sparse tree (\citet{Ye17jair}), 
 and on a shallow and 'thick' tree like the one generated by \citet{Lim20ijcai}. We provide full explanation how these trees are built in the caption of Table \ref{table:DESPOT_Lim__POMCP_table}.  The reported results in Table \ref{table:DESPOT_Lim__POMCP_table} are the mean time (in seconds) per planning session, each of horizon of $L$ steps, of our approach compared to calculating the objective using original rewards. We experiment with changing number of particles and planning horizon. 

Moreover, we consider two settings: 'I' and II'. Setting 'II', illustrated in Fig.~\ref{fig:plannig_simulation},  is more complex as the agent needs to move from bottom left corner to top right and the action space is \{left, right, up, down\} which creates symmetry. This symmetry makes it harder for the simplification to take effect and the agent needs to 're-simplify' a lot. Setting 'I' is easier: The agent needs to move from an initial point $x_0$ to the same height point $x_L$, which means the optimal path is just a straight line, while the action space is \{left, right\}. This easy setting allows the agent to utilize simplification to its full extent. Empty cells in Table \ref{table:DESPOT_Lim__POMCP_table} correspond to runs that planning session (for a regular objective) took longer than 35 [sec] and was stopped. Initial simplification level was set to $N^s = 0.1\cdot N$  and along the depth of the tree re-simplification had to take place from time to time (depending on the simulation). Specifically, whenever re-simplification was needed at some depth, a new simplification level was increased by factor of two, i.e. since $s_0=0.1$ then $s_1=0.2, s_2=0.4$  and so on until $s_n=1.0$.

It is clear from Table \ref{table:DESPOT_Lim__POMCP_table}, using our suggested simplification is a favorable approach, leading to speedup in all of the conducted experiments. Since the simplification bounds are analytical and used for eliminating branches in the belief tree of the original problem, the same optimal action is obtained with or without our simplification. In other words, we demonstrate a significant speedup while obtaining the same solution.

\begin{figure}[!ht]
	\centering
	\subfloat[ \label{fig:simplification_level_histogram_settingI}]{\includegraphics[width=0.42\columnwidth]{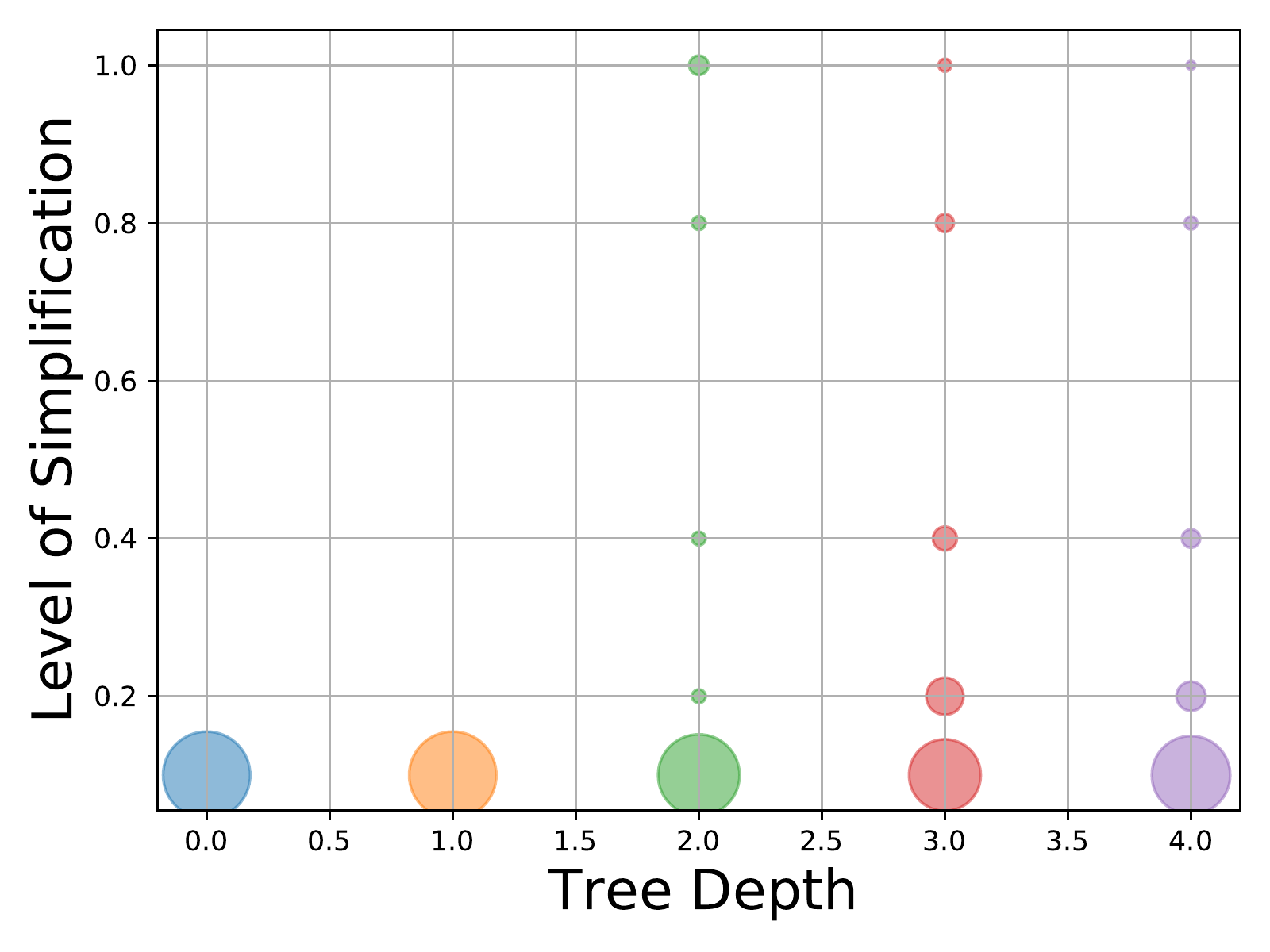}}
	\subfloat[ \label{fig:simplification_level_histogram_settingII}]{\includegraphics[width=0.42\columnwidth]{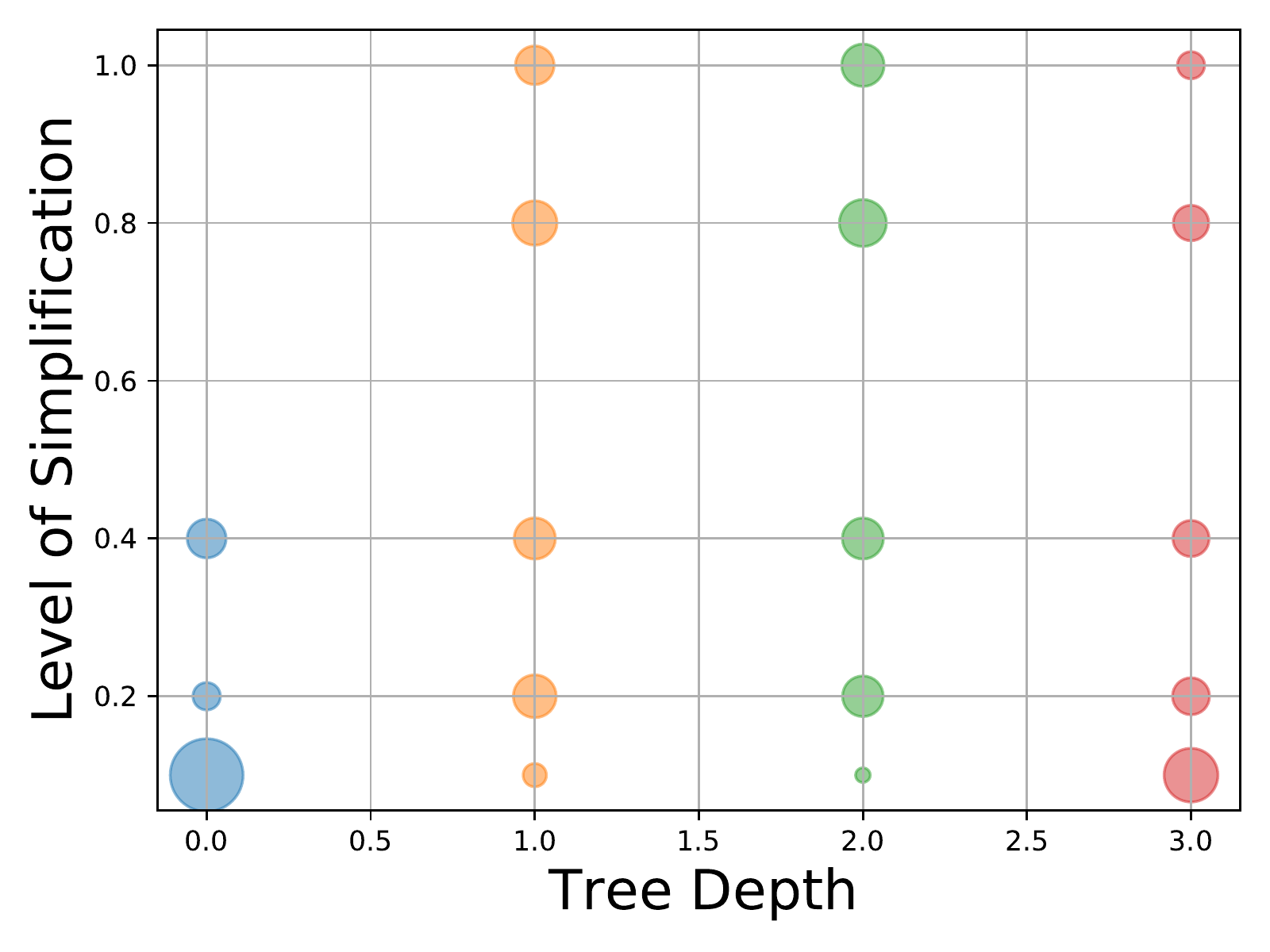}}
	\caption{Simplification level histogram illustration vs. tree depth for entire simulation (10 planning sessions). \textbf{(a)} Setting I using $N=50$ and Horizon of $4$. \textbf{(b)} Setting II using $N=20$ and Horizon of $3$. x-axis corresponding to tree nodes of some depth in the belief tree. y-axis corresponds to the simplification level needed to achieve pruning. The scale of the circles corresponds to how many nodes were in that category. Circles scales are normalized since the number of nodes grows exponentially going down the tree.}
	\label{fig:simplification_histogram}
\end{figure}

In Fig.~\ref{fig:simplification_histogram} we can get a glimpse into how our adaptive simplification performing in the tree depth for Settings I and II. The shown plots are essentially histograms that tell us what are the levels of simplification in the tree needed for pruning. It can be seen that indeed for Setting I the simplification is performing extremely well thus saving a lot of time. In the more difficult Setting II, indeed higher simplifications levels are more common. Nevertheless, as seen in Table \ref{table:DESPOT_Lim__POMCP_table}, we still get a significant speedup, while the speed-up for Setting I is even more drastic. 
This implies our simplification is able to identify by itself situations where we can save our resources (computation time) and all this without compromising on the accuracy of the desired solution.

%
%
%
%
%
%

\section{Conclusion}\label{sec:conclusion}
In this paper we have introduced SITH-BSP, an algorithmic paradigm able to speedup calculations when performing online POMDP planning, considering general distributions and belief-dependent rewards. The lion part of our approach such as the definition of simplification and LC bounds, is mathematically formulated in a general manner. It can be easily extended to many directions and may prove to be applicable for existing or future approaches. The more specific aspects of our approach (assuming Particle Filter for belief approximation) provides novel derivations for bounds over the differential entropy approximation and may be taken to other areas in the vast field of robotics. We have shown how the approach assumes an adaptive from, while re-using calculations and thus gives the exact optimal solution to the original problem in an effective manner.

In future work we would like to explore and identify more possible simplifications. It is also imperative we show how we can mount our existing simplification approach on existing approaches for online POMDP planning and by doing so show how general and beneficial our approach can be.

\section*{Acknowledgments}

This research was  supported by the Israel Science Foundation (ISF) and by 
a donation from the Zuckerman Fund to the Technion Center for Machine Learning and Intelligent Systems (MLIS).


\bibliographystyle{plainnat}
\bibliography{refs}

\begin{thebibliography}{13}
\providecommand{\natexlab}[1]{#1}
\providecommand{\url}[1]{\texttt{#1}}
\expandafter\ifx\csname urlstyle\endcsname\relax
  \providecommand{\doi}[1]{doi: #1}\else
  \providecommand{\doi}{doi: \begingroup \urlstyle{rm}\Url}\fi

\bibitem[{Boers} et~al.(2010){Boers}, {Driessen}, {Bagchi}, and
  {Mandal}]{Boers10fusion}
Y.~{Boers}, H.~{Driessen}, A.~{Bagchi}, and P.~{Mandal}.
\newblock Particle filter based entropy.
\newblock In \emph{2010 13th International Conference on Information Fusion},
  pages 1--8, 2010.
\newblock \doi{10.1109/ICIF.2010.5712013}.

\bibitem[Elimelech and Indelman(2018)]{Elimelech18ijrr_submitted}
Khen Elimelech and Vadim Indelman.
\newblock Simplified decision making in the belief space using belief
  sparsification.
\newblock \emph{Intl. J. of Robotics Research}, 12 2018.
\newblock Conditionally accepted.

\bibitem[Fehr et~al.(2018)Fehr, Buffet, Thomas, and Dibangoye]{Fehr18nips}
Mathieu Fehr, Olivier Buffet, Vincent Thomas, and Jilles Dibangoye.
\newblock rho-pomdps have lipschitz-continuous epsilon-optimal value functions.
\newblock In S.~Bengio, H.~Wallach, H.~Larochelle, K.~Grauman, N.~Cesa-Bianchi,
  and R.~Garnett, editors, \emph{Advances in Neural Information Processing
  Systems 31}, pages 6933--6943. Curran Associates, Inc., 2018.

\bibitem[Fischer and Tas(2020)]{Fischer20icml}
Johannes Fischer and Omer~Sahin Tas.
\newblock Information particle filter tree: An online algorithm for pomdps with
  belief-based rewards on continuous domains.
\newblock In \emph{Intl. Conf. on Machine Learning (ICML)}, Vienna, Austria,
  2020.

\bibitem[Garg et~al.(2019)Garg, Hsu, and Lee]{Garg19rss}
Neha~P Garg, David Hsu, and Wee~Sun Lee.
\newblock Despot-$\alpha$: Online pomdp planning with large state and
  observation spaces.
\newblock In \emph{Robotics: Science and Systems (RSS)}, 2019.

\bibitem[Hoerger et~al.(2019)Hoerger, Kurniawati, and Elfes]{Hoerger19isrr}
Marcus Hoerger, Hanna Kurniawati, and Alberto Elfes.
\newblock Multilevel monte-carlo for solving pomdps online.
\newblock In \emph{Proc. International Symposium on Robotics Research (ISRR)},
  2019.

\bibitem[Lim et~al.(2020)Lim, Tomlin, and Sunberg]{Lim20ijcai}
Michael~H. Lim, Claire Tomlin, and Zachary~N. Sunberg.
\newblock Sparse tree search optimality guarantees in pomdps with continuous
  observation spaces.
\newblock In \emph{Intl. Joint Conf. on AI (IJCAI)}, pages 4135--4142, 7 2020.

\bibitem[Papadimitriou and Tsitsiklis(1987)]{Papadimitriou87math}
C.~Papadimitriou and J.~Tsitsiklis.
\newblock The complexity of markov decision processes.
\newblock \emph{Mathematics of operations research}, 12\penalty0 (3):\penalty0
  441--450, 1987.

\bibitem[Silver and Veness(2010)]{Silver10nips}
David Silver and Joel Veness.
\newblock Monte-carlo planning in large pomdps.
\newblock In \emph{Advances in Neural Information Processing Systems (NIPS)},
  pages 2164--2172, 2010.

\bibitem[Smith and Simmons(2004)]{Smith04uai}
T.~Smith and R.~Simmons.
\newblock Heuristic search value iteration for pomdps.
\newblock In \emph{Conf. on Uncertainty in Artificial Intelligence (UAI)},
  pages 520--527, 2004.

\bibitem[Smith and Simmons(2005)]{Smith05uai}
T.~Smith and R.~Simmons.
\newblock Point-based pomdp algorithms: Improved analysis and implementation.
\newblock In \emph{Conf. on Uncertainty in Artificial Intelligence (UAI)},
  pages 542--547, 2005.

\bibitem[Sunberg and Kochenderfer(2018)]{Sunberg18icaps}
Zachary Sunberg and Mykel Kochenderfer.
\newblock Online algorithms for pomdps with continuous state, action, and
  observation spaces.
\newblock In \emph{Proceedings of the International Conference on Automated
  Planning and Scheduling}, volume~28, 2018.

\bibitem[Ye et~al.(2017)Ye, Somani, Hsu, and Lee]{Ye17jair}
Nan Ye, Adhiraj Somani, David Hsu, and Wee~Sun Lee.
\newblock Despot: Online pomdp planning with regularization.
\newblock \emph{JAIR}, 58:\penalty0 231--266, 2017.

\end{thebibliography}

\clearpage
\onecolumn
	\section{Appendix}
	\subsection{Proof for Theorem \ref{theorem:LC_bounds}}\label{proof:theorem_1}
	\begin{equation}\label{eq:LC_upper_bound}
	\begin{split}
	&J(b_i,\pi_{i:i+L-1})= \expt{z_{i+1:i+L}}{\sum_{j=0}^{L-1}r(b_{i+j},a_{i+j}) \mid H_i}=
	r(b_i,a_i)+\int \prob{z_{i+1}\mid H_i, a_i}[\ r(b_{i+1},a_{i+1})+\\
	&\int \prob{z_{i+2}\mid H_i, a_{i+1}, z_{i+1}} [\ r(b_{i+2},a_{i+2})+
	\int \cdots[]dz_{i+L} \cdots]\ dz_{i+2} ]\ dz_{i+1} \maxim{\leq}{L.C.}\\
	&r(b_i^s, a_i)+ \lambda_r \cdot d(b_i, b_i^s)+ \int \prob{z_{i+1}\mid H_i, a_i}
	[\ r(b_{i+1}^s,a_{i+1})+ \lambda_r \cdot d(b_{i+1},b_{i+1}^s) + \\ 
	&\int \prob{z_{i+2}\mid H_i, a_{i+1}, z_{i+1}}[\ r(b_{i+2}^s,a_{i+2})+
	\lambda_r \cdot d(b_{i+2},b_{i+2}^s) +\int [\cdots]dz_{i+L}] \cdots]\ dz_{i+2} ]\ dz_{i+1}=
	\mathcal{UB}(b_i, \pi_{i+})
	\end{split}
	\end{equation}
	
	The lower bound is calculated in the exact same manner using the other side of \ref{eq:LC_rewards}.
	\begin{equation}\label{eq:LC_lower_bound}
	\begin{split}
	&J(b_i,\pi_{i:i+L-1})\maxim{\geq}{L.C.}
	r(b_i^s, a_i)- \lambda_r \cdot d(b_i, b_i^s)+ \int \prob{z_{i+1}\mid H_i, a_i}
	[\ r(b_{i+1}^s,a_{i+1})- \lambda_r \cdot d(b_{i+1},b_{i+1}^s) + \\
	&\int \prob{z_{i+2}\mid H_i, a_{i+1}, z_{i+1}}[\ r(b_{i+2}^s,a_{i+2})-
	\lambda_r \cdot d(b_{i+2},b_{i+2}^s) +\int [\cdots]dz_{i+L}] \cdots]\ dz_{i+2} ]\ dz_{i+1}=
	\mathcal{LB}(b_i, \pi_{i+})
	\end{split}
	\end{equation}

	\subsection{Proof for Theorem \ref{theorem:lower_upper_I}}\label{proof:theorem_2}
	\begin{equation}\label{eq:lower_I}
	\begin{split}
	&\log \left[\sum_i \prob{z_{k+1}\mid x_{k+1}^i} w_k^i\right]=
	\log \left[\sum_{i\in A^s_{k+1}} \prob{z_{k+1}\mid x_{k+1}^i} w_k^i+ 
	\sum_{i\in \neg A^s_{k+1}} \prob{z_{k+1}\mid x_{k+1}^i} w_k^i\right]\\
	&=\log \left[\sum_{i\in A^s_{k+1}} \prob{z_{k+1}\mid x_{k+1}^i} w_k^i \right]+
	\log \left[1+ \frac{\sum_{i\in \neg A^s_{k+1}} \prob{z_{k+1}\mid x_{k+1}^i} w_k^i}{\sum_{i\in A^s_{k+1}} \prob{z_{k+1}\mid x_{k+1}^i} w_k^i} \right] \geq
	\log \left[\sum_{i\in A^s_{k+1}} \prob{z_{k+1}\mid x_{k+1}^i} w_k^i \right]
	\end{split}
	\end{equation}
	
	We upper bound the first term by bounding the later 'log' term in the derivation of \ref{eq:lower_I} just before the inequity.
	\begin{equation}\label{eq:upper_I}
	\begin{split}
	&\log \left[1+ \frac{\sum_{i\in \neg A^s_{k+1}} \prob{z_{k+1}\mid x_{k+1}^i} w_k^i}{\sum_{i\in A^s_{k+1}} \prob{z_{k+1}\mid x_{k+1}^i} w_k^i} \right] \leq 
	\log \left[1+ \frac{\sum_{i\in \neg A^s_{k+1}} n\cdot w_k^i}{\sum_{i\in A^s_{k+1}} \prob{z_{k+1}\mid x_{k+1}^i} w_k^i} \right] \leq\\
	&\log \left[1+ \frac{n\cdot \left[1-\sum_{i\in A^s_{k+1}}w_{k}^i\right]}{\sum_{i\in A^s_{k+1}} \prob{z_{k+1}\mid x_{k+1}^i} w_k^i} \right]
	\end{split}
	\end{equation}

%
	\subsection{Proof for Theorem \ref{theorem:lower_upper_II}}\label{proof:theorem_3}
	\begin{equation}\label{eq:lower_II}
	\begin{split}
	&-\sum_i w_{k+1}^i\cdot\log\left[\prob{z_{k+1}\mid x_{k+1}^i} \sum_j\prob{x_{k+1}^i\mid x_k^j, a_k}w_k^j\right]=\\
	&-\sum_{i \in A^s_{k+1}} w_{k+1}^i\cdot\log\left[\prob{z_{k+1}\mid x_{k+1}^i} \sum_j\prob{x_{k+1}^i\mid x_k^j, a_k}w_k^j\right]-
	\sum_{i \in \neg A^s_{k+1}} w_{k+1}^i\cdot\log\left[\prob{z_{k+1}\mid x_{k+1}^i} \sum_j \prob{x_{k+1}^i\mid x_k^j, a_k}w_k^j\right] \\
	&\geq -\sum_{i \in A^s_{k+1}} w_{k+1}^i\cdot\log\left[\prob{z_{k+1}\mid x_{k+1}^i} \sum_j\prob{x_{k+1}^i\mid x_k^j, a_k}w_k^j\right]
	-
	\sum_{i \in \neg A^s_{k+1}} w_{k+1}^i\cdot\log\left[ m \cdot \prob{z_{k+1}\mid x_{k+1}^i} \cdot \sum_j w_k^j\right]=\\
	&-\sum_{i \in A^s_{k+1}} w_{k+1}^i\cdot\log\left[\prob{z_{k+1}\mid x_{k+1}^i} \sum_j\prob{x_{k+1}^i\mid x_k^j, a_k}w_k^j\right]
	-
	\sum_{i \in \neg A^s_{k+1}} w_{k+1}^i\cdot\log\left[ m \cdot \prob{z_{k+1}\mid x_{k+1}^i}\right]
	\end{split}
	\end{equation}
	
	\begin{equation}\label{eq:upper_II}
	\begin{split}
	&-\sum_i w_{k+1}^i\cdot\log\left[\prob{z_{k+1}\mid x_{k+1}^i} \sum_j\prob{x_{k+1}^i\mid x_k^j, a_k}w_k^j\right]=
	-\sum_i w_{k+1}^i\cdot\log\left[\sum_{j \in A^s_k} \prob{z_{k+1}\mid x_{k+1}^i} \prob{x_{k+1}^i \mid x_k^j, a_k}w_k^j\right] -\\
	&-\sum_i w_{k+1}^i\cdot\log\left[1+ \frac{\sum_{j \in \neg A^s_k} \prob{z_{k+1}\mid x_{k+1}^i} \prob{x_{k+1}^i \mid x_k^j, a_k}w_k^j}{\sum_{j \in A^s_k} \prob{z_{k+1}\mid x_{k+1}^i} \prob{x_{k+1}^i \mid x_k^j, a_k}w_k^j}\right]
	\leq -\sum_i w_{k+1}^i\cdot\log\left[\sum_{j \in A^s_k} \prob{z_{k+1}\mid x_{k+1}^i} \prob{x_{k+1}^i \mid x_k^j, a_k}w_k^j\right]
	\end{split}
	\end{equation}

\subsection{Pruning tree branches using reward bounds}\label{appen:branch_and_bound}
Usually when planning into the future a planning tree, or a belief tree in the more general case, is built in some manner. This tree approximates the expectation of cumulative future rewards given different possible policies. In order to decide which action should be taken at the root of the tree, rewards should be summed bottom up (leafs to root). This weighted summation for the different routes in the tree, is nothing but the objective function \eqref{eq:objective}. Once the rewards are propagated up the tree, the action (at the root) that present greater future cumulative reward should be chosen, i.e. choose the most promising subtree of the original tree (illustration in Fig.~\ref{fig:action_graph}). Due to the recursive nature of \eqref{eq:objective}, \eqref{eq:bellman} this formulation is also recursive and is applied in each belief node of the belief tree. I.e., in each node we propagate up the action that has the biggest corresponding subtree cumulative reward. Thus we get the optimal policy.

A possible way to improve this setting is bounding the tree branches. Meaning, each belief node $b_{k+j}$ in the belief tree  has children subtrees corresponding to the different actions that can be taken from $b_{k+j}$. Each child subtree has it's own upper and lower bound $\{\mathcal{LB}^m, \mathcal{UB}^m\}_{m=1}^{|\mathcal{A}|}$ that we somehow got. So, according to the bounds, when some actions (subtrees) seem to be less promising than their sibling action, we can avoid expanding this tree branch in the first place. Though this approach is sub optimal according to \citet{Lim20ijcai}. An alternative way to speedup the process is eliminating existing branches (subtrees or actions) according to these bounds. It becomes possible when for two sibling subtrees $m',m''$ corresponding to two different actions, we get $\mathcal{LB}_{m'} > \mathcal{UB}_{m''}$ or $\mathcal{LB}_{m''} > \mathcal{UB}_{m'}$. E.g., in Fig.~\ref{fig:action_graph_s1} the lower bound of $\pi''''$ is higher than all other actions upper bounds. However this becomes problematic if (a) the bounds are not cheaper to calculate than the original objective of some tree. (b) We cannot eliminate all actions but one since the bounds are not tight enough. E.g.~in Fig.~\ref{fig:action_graph_s0} one cannot say for sure that policy $\pi''''$ is better than policy $\pi''$ since the latter upper bound is higher than the former lower bound.

\subsection{Adaptive Simplification Illustrative Example}\label{appen:adaptive_simulation}
Consider Fig.~\ref{fig:adaptive_bounds} and  assume the subtrees to $b_i^1$ were solved using simplification levels that hold $s^2=s^1+1, s^2<s^3,s^4$. Further assume the immediate reward simplification is $s=s^1$. According to definitions above this means that for $b_i^1$,  $s^{j=1}=\min\{s^1, s^{l=1}, s^{l=2}\}$ and $s^{j=2}=\min\{s^1, s^{l=3}, s^{l=4}\}$. Now, we consider the case the existing bounds of the subtrees were not tight enough to prune, we adapt simplification level of the tree starting from $b_i^1:s^1\rightarrow s^1+1$. Since $s^1 < s^1+1$ we re-simplify the subtree corresponding to simplification level of $s^1$ to simplification level $s^1+1$, i.e.~to a finer simplification.

However we do not need to re-simplify subtrees corresponding to $s^2,s^3,s^4$: The tree corresponding to  $s^2$ is already simplified to the currently desired level thus we can use its existing bounds. For the two other trees, their current simplification levels, $s^3$ and $s^4$, are higher (finer) than the desired $s^1+1$ level, and since the bounds are tighter as simplification level increases we can use their existing tighter bounds without the need to 'go-back' to a coarser level of simplification. If we can now prune one of the actions, we keep pruning up the tree. If pruning is still not possible, we need to adapt simplification again with simplification level $s^1+2$.

\subsection{Lipschitz Continuity}\label{appen:LC_definition}
The Lipschitz continuity property (denoted as LC) of a function means that given two metric spaces $(X,d_X), (Y,d_Y)$, where  $d_X$ denotes the metric on set $X$ and $d_Y$ is the metric on set $Y$, a function $f:X \rightarrow Y$ is Lipschitz continuous if there exists a real constant $K\geq 0$ such that $\forall x_1, x_2 \in X: d_Y(f(x_1),f(x_2)) \leq K\cdot d_X(x_1, x_2)$.

\newpage 
\subsection{Algorithms}
\begin{algorithm}
	\caption{Prune Branches}
	\begin{algorithmic}[1]
		\Procedure{Prune}{}\\
		\hspace*{\algorithmicindent} \textbf{Input:} (belief-tree root, $b$; bounds of root's children, $\{\mathcal{LB}^m, \mathcal{UB}^m\}_{m=1}^{C}$)\Comment{$C$ is the number of child branches going out of $b$.}
		\State $\mathcal{LB}^{\star} \leftarrow \maxim{max}{m} \{\mathcal{LB}^m\}_{m=1}^{C}$
		\ForAll{children of $b$}
		\If{$\mathcal{LB}^{\star} > \mathcal{UB}^m$}
		\State prune child $m$ from the belief tree
		\EndIf
		\EndFor
		\EndProcedure
	\end{algorithmic}
	\label{alg:prune}
\end{algorithm}

\begin{algorithm}
	\caption{Simplified Information Theoretic Belief Space Planning (SITH-BSP)}
	\begin{algorithmic}[1]
		\Procedure{Find Optimal Policy}{belief-tree: $\mathbb{T}$}
		\State $s\leftarrow s_0$
		\State \textbf{return} \Call{Adapt Simplification}{$\mathbb{T}$,$s$}
		\EndProcedure
		
		\Procedure{Adapt Simplification}{belief-tree: $\mathbb{T}, s_i$}
		\If { $\mathbb{T}$ is a leaf}
		\State \textbf{return} $\{\mathbf{lb}, \mathbf{ub}\}$ \Comment{Corresponds to immediate reward bounds over the leaf \eqref{eq:bound_reward_general}.}
		\EndIf
		\State Set simplification level: $s\leftarrow s_i$
		\ForAll{subtrees  $\mathbb{T}'$ in $\mathbb{T}$}
		\State \Call{Adapt Simplification}{$\mathbb{T}'$,$s$}
		\State Calculate $\mathcal{LB}^{s^j}, \mathcal{UB}^{s^j}$ according to $s$ and \eqref{eq:bellman_bounds_simp_level}
		\EndFor
		\State Using $\{\mathcal{LB}^{s^j}, \mathcal{UB}^{s^j}\}_{j=1}^{|\mathcal{A}|}$ and Alg. \ref{alg:prune} prune branches
		\While{not all $\mathbb{T}'$ but 1 in $\mathbb{T}$ pruned}
		\State Increase simplification level: $s\leftarrow s+1$
		\State \Call{Adapt Simplification}{$\mathbb{T}$,$s$}
		\EndWhile
		\State Update $\{\mathcal{LB}^{s^j \star}, \mathcal{UB}^{s^j \star}\}$ according to \eqref{eq:bellman_bounds_simp_level_optimal}
		\State \textbf{return} optimal action branch that left $a^{\star}$  and $\{\mathcal{LB}^{s^j \star}, \mathcal{UB}^{s^j \star}\}$. 
		\EndProcedure
	\end{algorithmic}
	\label{alg:simp_planning}
\end{algorithm}

\onecolumn

\subsection{Additional Entropy Results}
\begin{figure*}[th]
	\centering
	\includegraphics[width=\columnwidth]{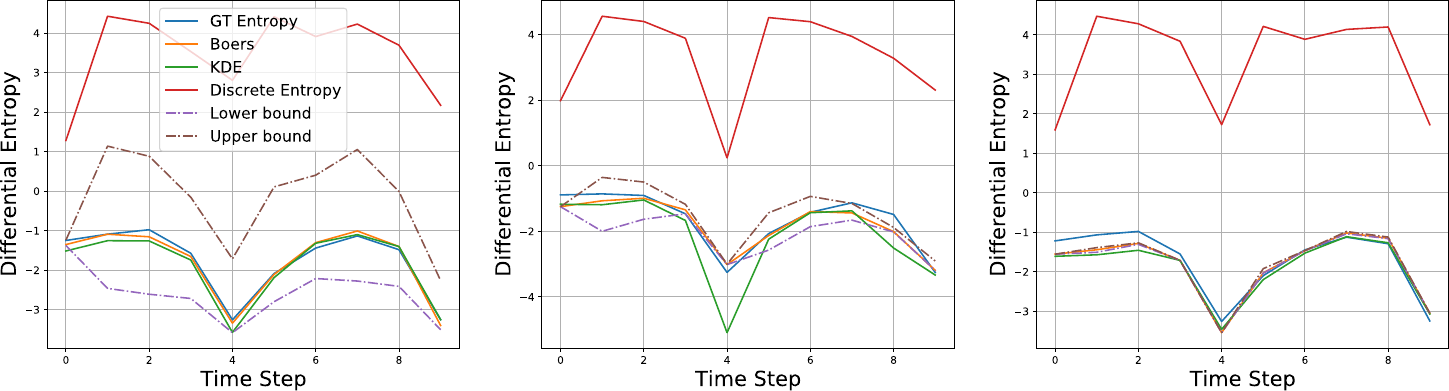}
	\caption{Differential Entropy Approximations ans Bounds. Calculations were done using $100$ particles. From left to right: Simplification is $N^s=\{0.1, 0.5, 0.9\}\cdot N$}
	\label{fig:ent_bounds_100}
\end{figure*}

\begin{figure*}[th]
	\centering
	\includegraphics[width=\columnwidth]{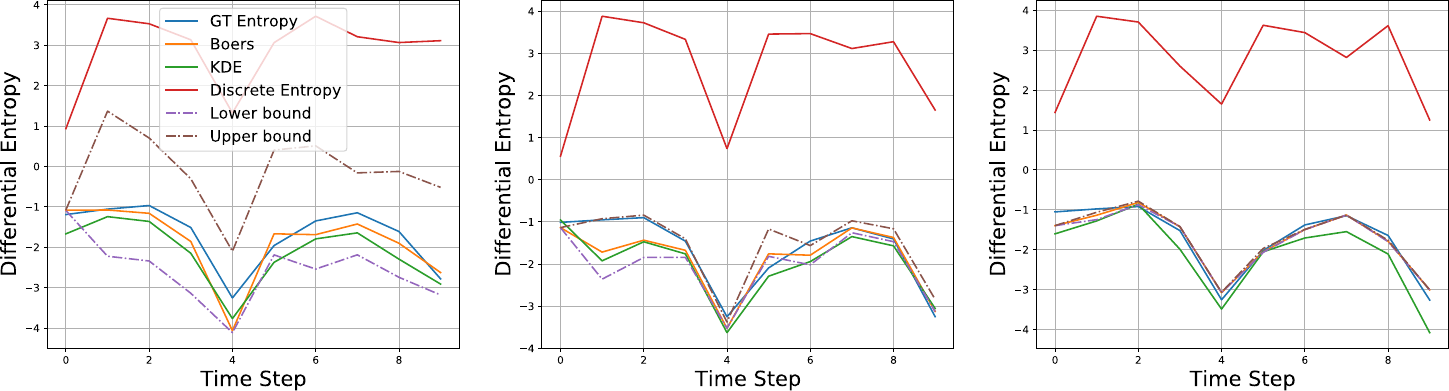}
	\caption{Differential Entropy Approximations ans Bounds. Calculations were done using $50$ particles. From left to right: Simplification is $N^s=\{0.1, 0.5, 0.9\}\cdot N$}
	\label{fig:ent_bounds_50}
\end{figure*}

\begin{figure*}[th]
	\centering
	\includegraphics[width=\columnwidth]{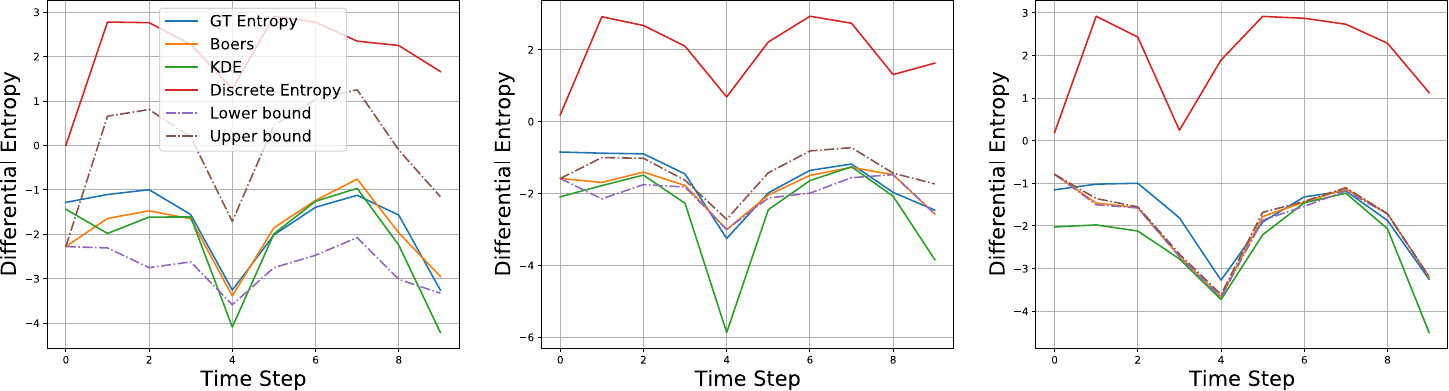}
	\caption{Differential Entropy Approximations ans Bounds. Calculations were done using $20$ particles. From left to right: Simplification is $N^s=\{0.1, 0.5, 0.9\}\cdot N$}
	\label{fig:ent_bounds_20}
\end{figure*}

\label{sec:supplementary}
\end{document}